\documentclass[conference]{IEEEtran}
\IEEEoverridecommandlockouts
\usepackage{cite}
\usepackage{amsmath,amssymb,amsfonts}
\usepackage{algorithmic}
\usepackage{graphicx}
\usepackage{textcomp}
\usepackage{xcolor}
\usepackage[caption=false,font=small,labelfont=rm,textfont=rm]{subfig}
\usepackage{booktabs}
\usepackage{multirow}
\usepackage[shortlabels]{enumitem}
\usepackage[normalem]{ulem}
\useunder{\uline}{\ul}{}
\usepackage{algorithm}
\usepackage{url}
\usepackage{longtable}

\def\BibTeX{{\rm B\kern-.05em{\sc i\kern-.025em b}\kern-.08em
    T\kern-.1667em\lower.7ex\hbox{E}\kern-.125emX}}
\begin{document}

\title{SMPLX-Lite: A Realistic and Drivable Avatar Benchmark with Rich Geometry and Texture Annotations\\

}

\author{Yujiao Jiang$^{1}$ $\quad$
        Qingmin Liao$^{1}$ $\quad$
        Zhaolong Wang$^{1,2}\quad$
        Xiangru Lin$^{2}\quad$ \\
        Zongqing Lu$^{1}\quad$ 
        Yuxi Zhao$^{1}\quad$ 
        Hanqing Wei$^{3}\quad$ 
        Jingrui Ye$^{1}\quad$ 
        Yu Zhang$^{2}\quad$ 
        Zhijing Shao$^{2,4*}$  \\ 
        $^1$Shenzhen International Graduate School, Tsinghua University$\quad$
        $^2$Prometheus Vision Technology Co., Ltd.\\
        $^3$Beijing University of Aeronautics and Astronautics \\
        $^4$The Hong Kong University of Science and Technology (Guangzhou)\\
{\tt\small jiangyj20@mails.tsinghua.edu.cn, neil.szj@prometh.xyz}
}


\maketitle

\begin{abstract}
Recovering photorealistic and drivable full-body avatars is crucial for numerous applications, including virtual reality, 3D games, and tele-presence.
Most methods, whether reconstruction or generation, require large numbers of human motion sequences and corresponding textured meshes.
To easily learn a drivable avatar, a reasonable parametric body model with unified topology is paramount.
However, existing human body datasets either have images or textured models and lack parametric models which fit clothes well. 
We propose a new parametric model \textbf{SMPLX-Lite-D}, which can fit detailed geometry of the scanned mesh while maintaining stable geometry in the face, hand and foot regions.
We present \textbf{SMPLX-Lite} dataset, the most comprehensive clothing avatar dataset with multi-view RGB sequences, keypoints annotations, textured scanned meshes, and textured SMPLX-Lite-D models.
With the SMPLX-Lite dataset, we train a conditional variational autoencoder model that takes human pose and facial keypoints as input, and generates a photorealistic drivable human avatar. 
\end{abstract}


\section{Introduction}
\label{sec:intro}
\begin{figure*}[htbp]
	\centering
	\subfloat[Color Image]{\label{fig:Color Image} \includegraphics[width=0.136\textwidth]{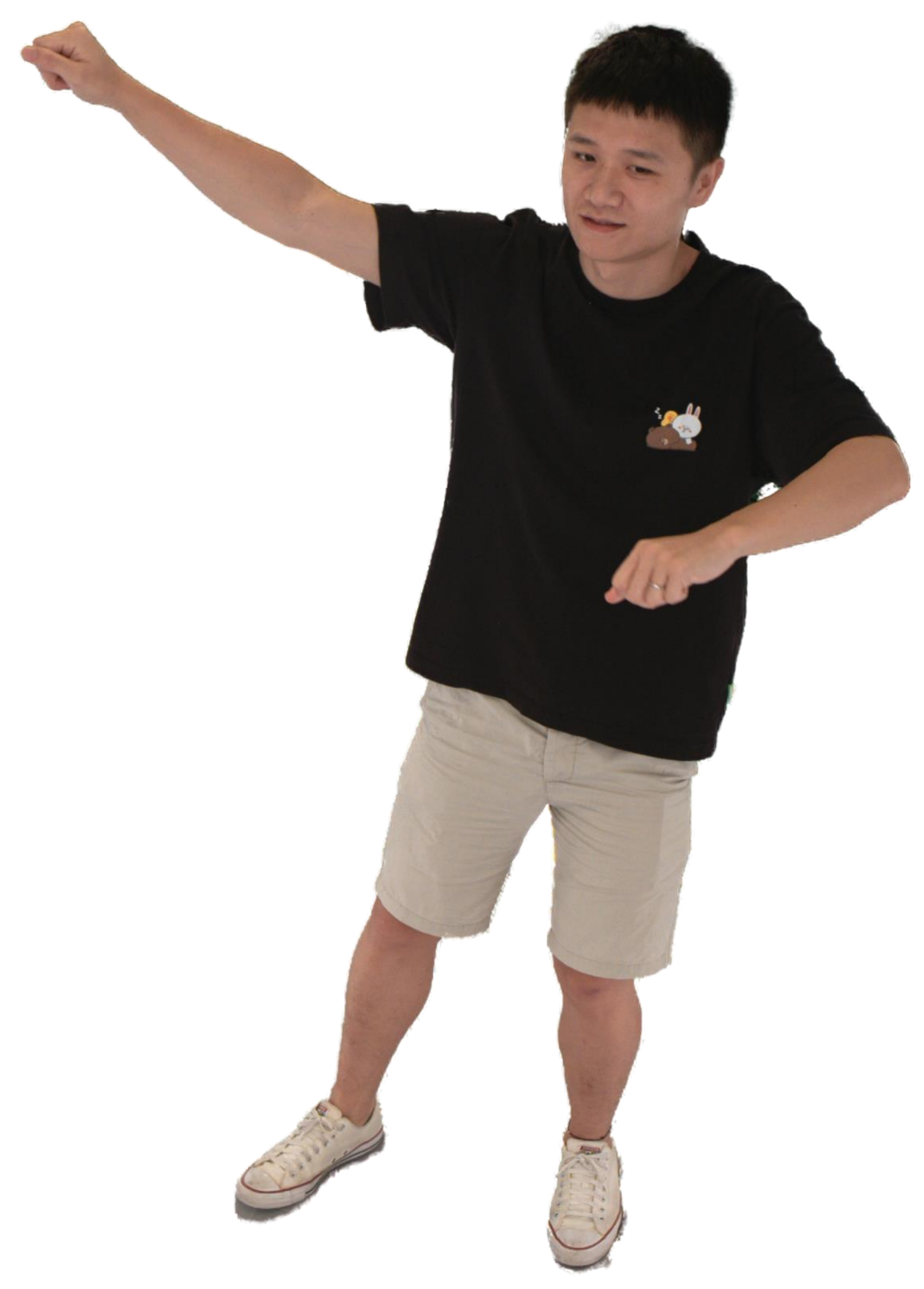}}
	\subfloat[Keypoints]{\label{fig:Keypoints} \includegraphics[width=0.136\textwidth]{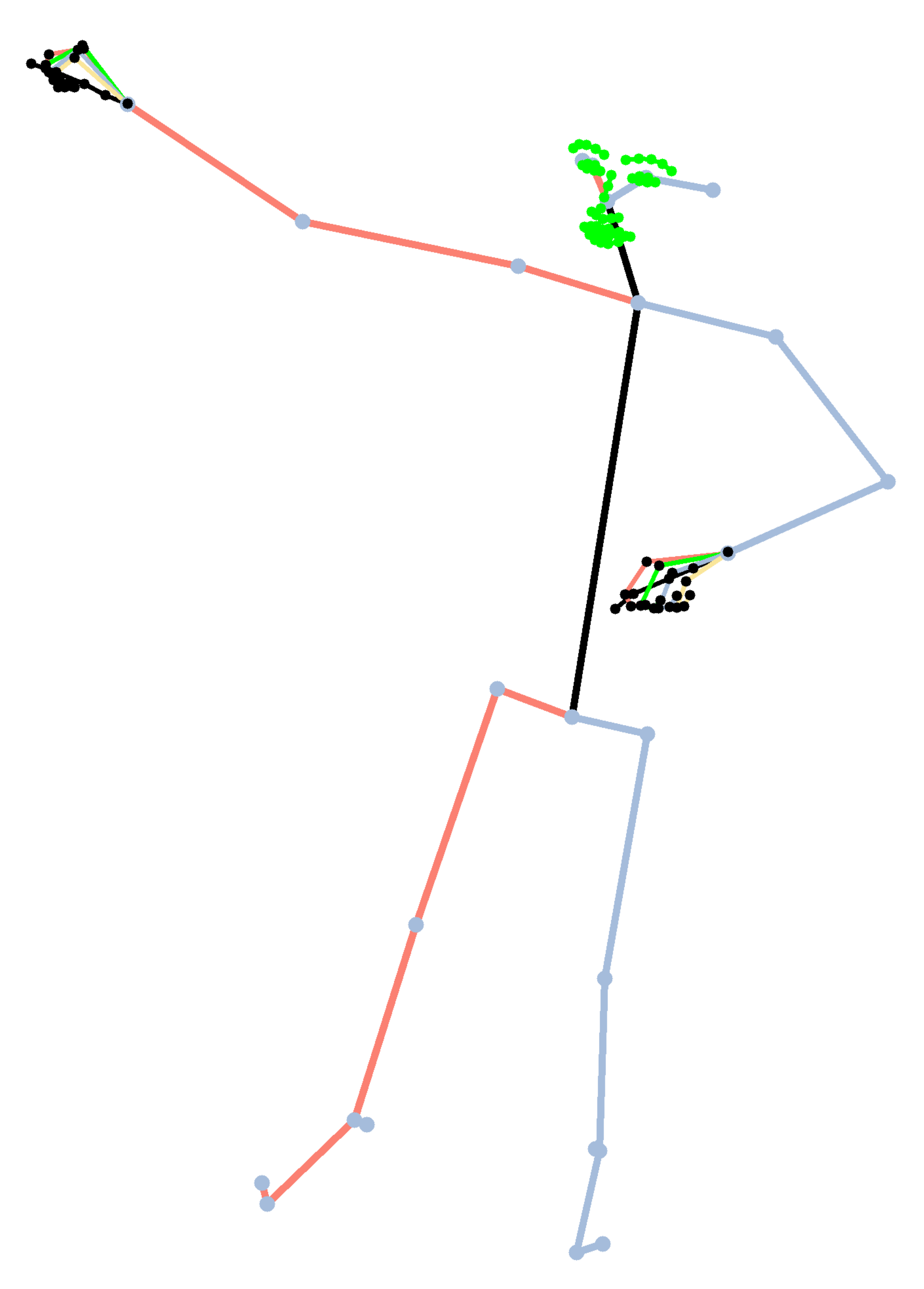}}
	\subfloat[SMPL-X]{\label{fig:SMPL-X} \includegraphics[width=0.136\textwidth]{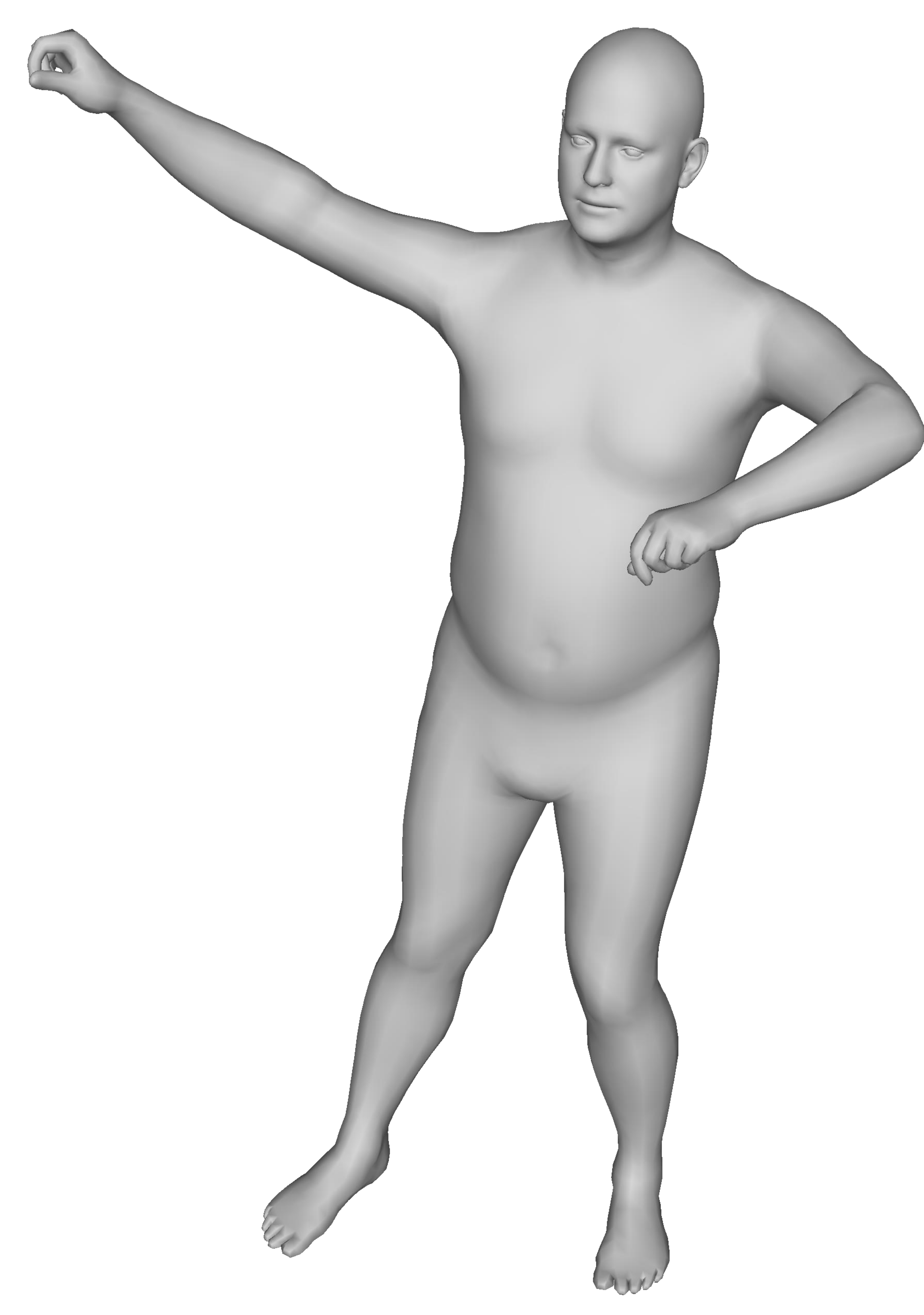}}
	\subfloat[Scanned Mesh]{\label{fig:Scanned Mesh} \includegraphics[width=0.136\textwidth]{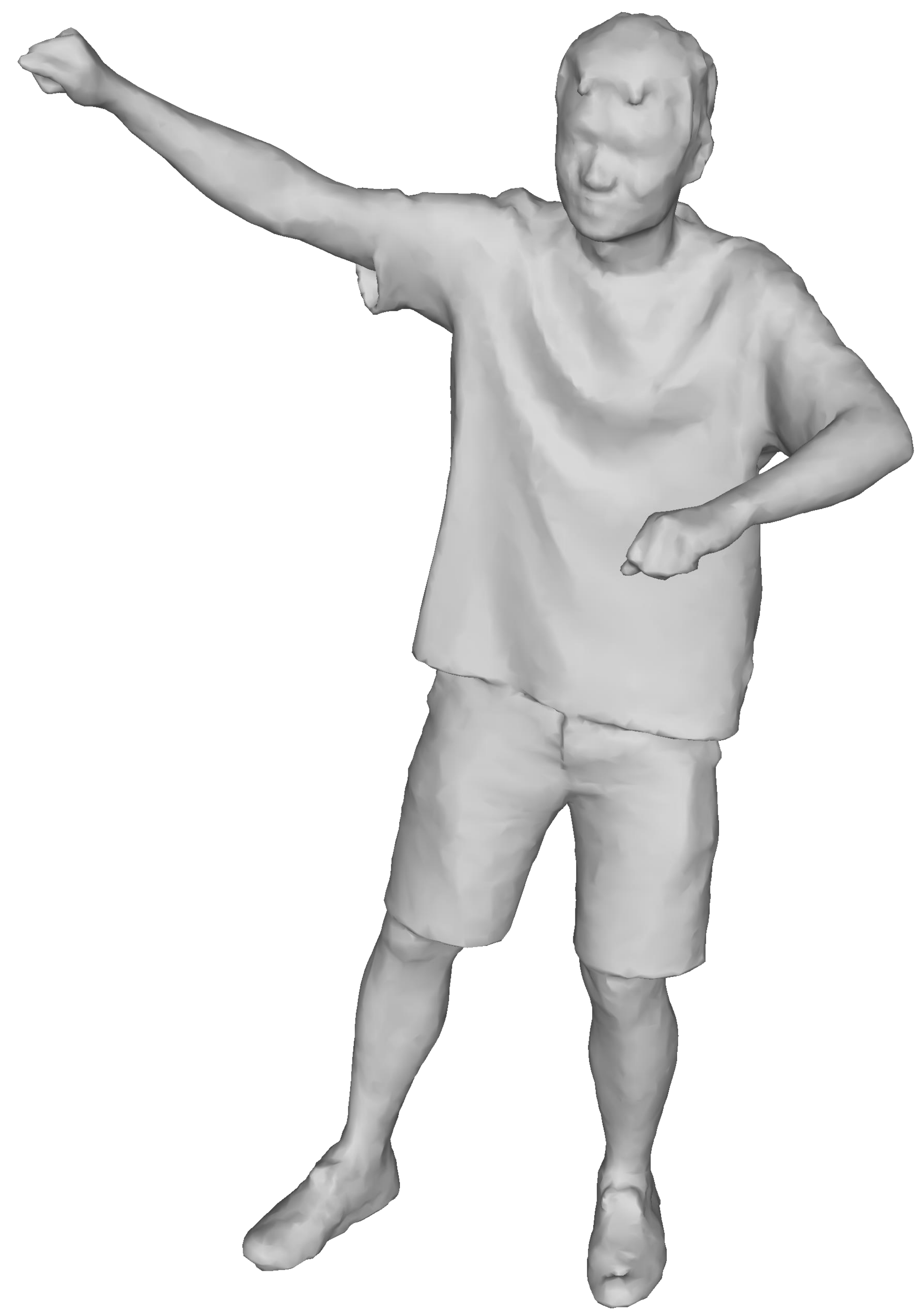}}
	\subfloat[Scanned Texture]{\label{fig:Scanned Texture} \includegraphics[width=0.136\textwidth]{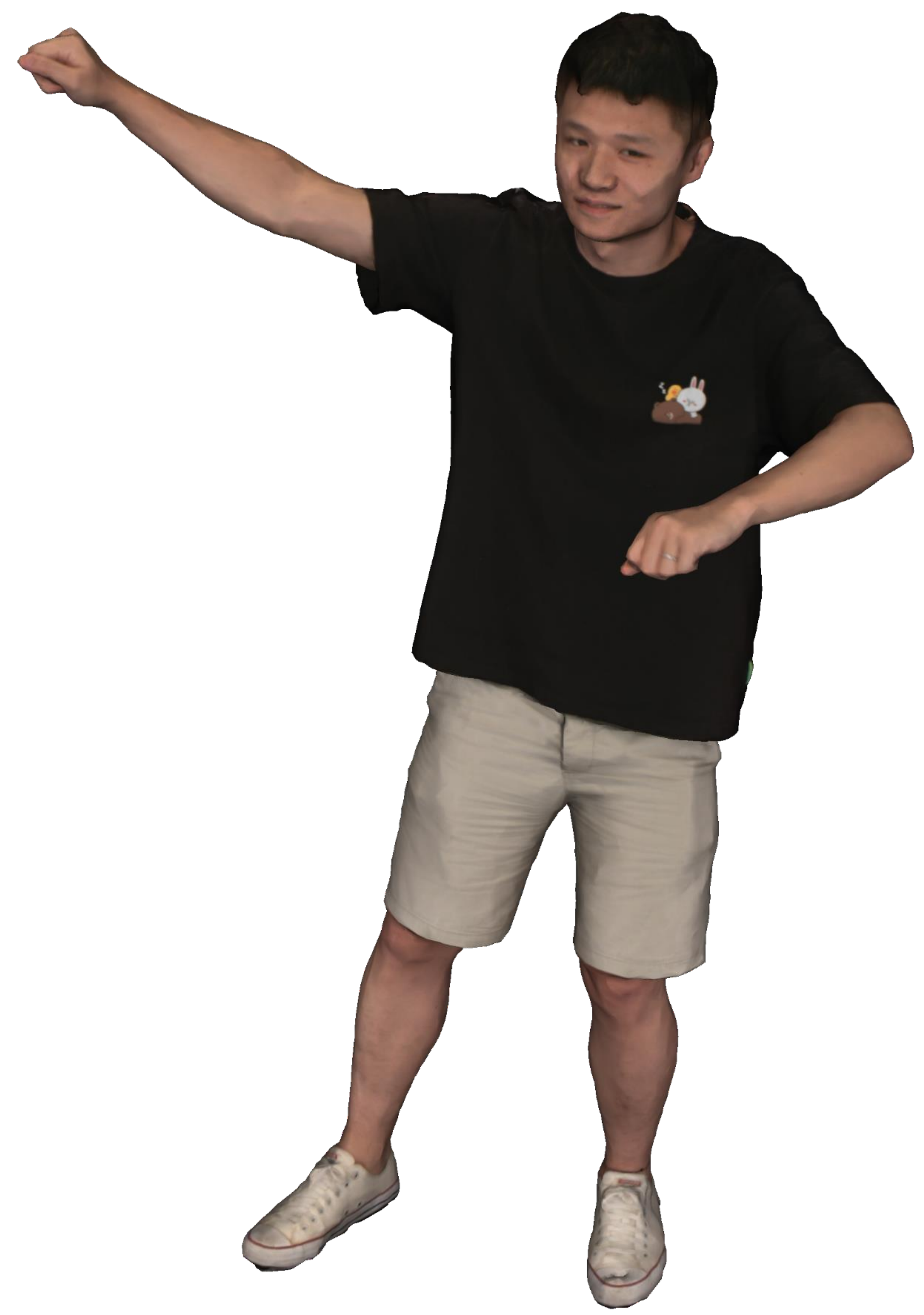}}
	\subfloat[Lite-D]{\label{fig:Lite-D} \includegraphics[width=0.136\textwidth]{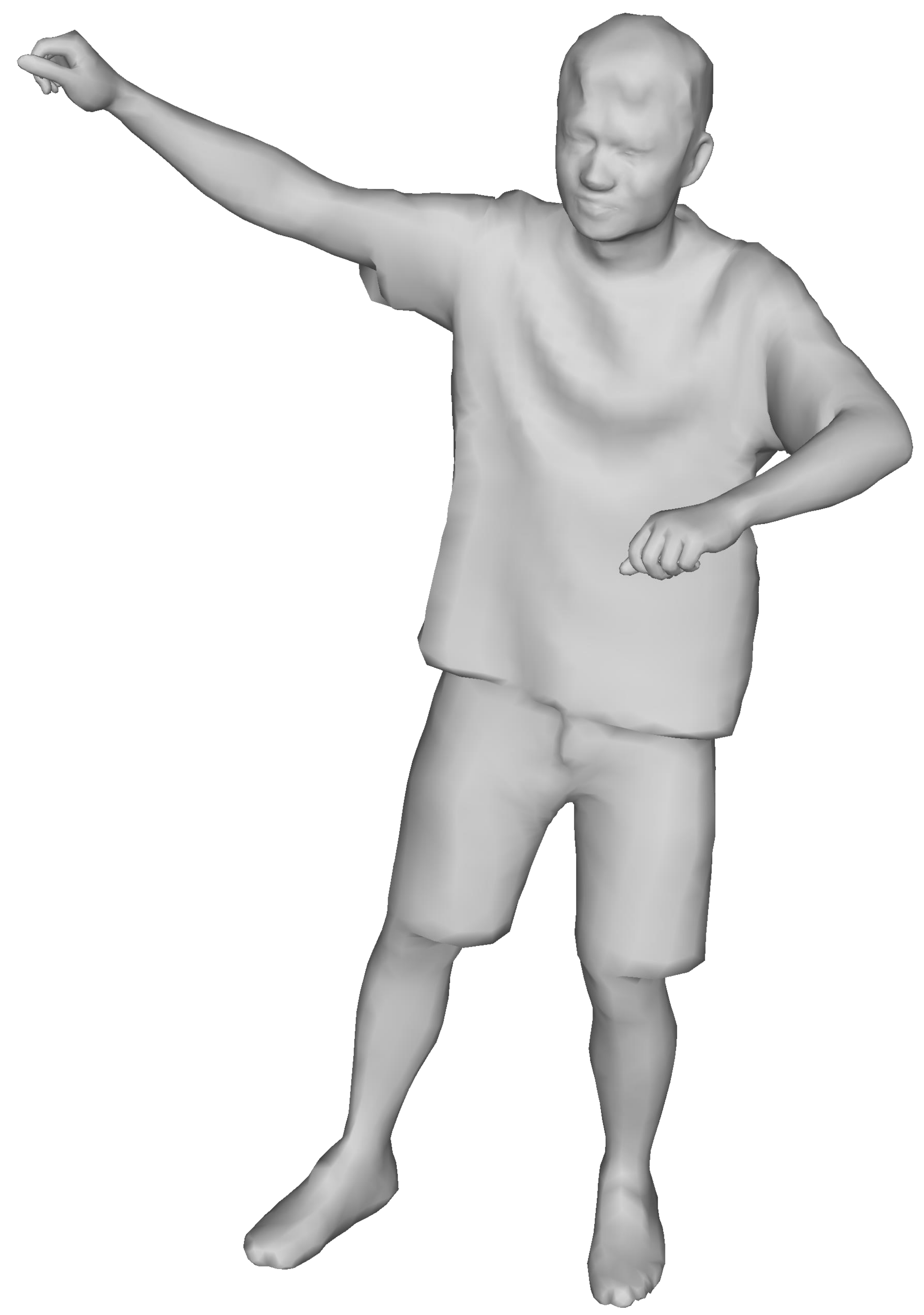}}
	\subfloat[Lite-D Texture]{\label{fig:Lite-D Texture} \includegraphics[width=0.136\textwidth]{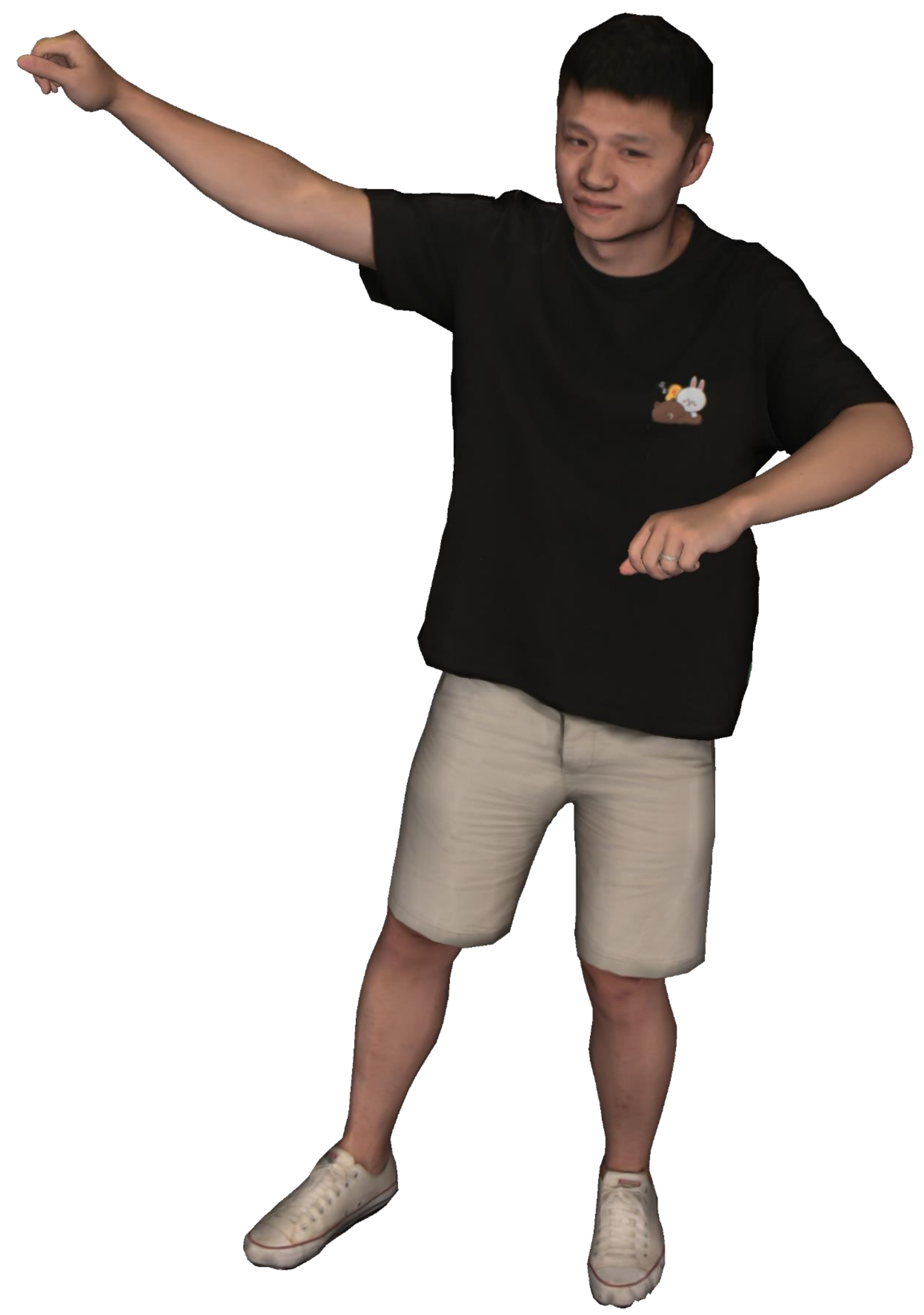}}
	
	\caption{SMPLX-Lite dataset has multiple types of data formats and annotations and is the most comprehensive dataset currently available in the drivable avatar area. We demonstrate a) color image, b) keypoints, c) SMPL-X registration, d) scanned mesh with e) scanned texture, and f) SMPLX-Lite-D mesh with g) SMPLX-Lite-D texture.}
	\label{fig:All data}
\end{figure*}

Existing methods generally reconstruct clothed human models from images or videos. 
One class of methods based on the neural radiance field\cite{2020PIFu, 2020PIFuHD, Dong_2022_PINA} utilizes an implicit functional representation\cite{chen2019learning, 2019DeepSDF} that allows pixel-level alignment with the image, but lacks an explicit geometric representation.

Another class of methods is based on parametric models (e.g., SMPL\cite{SMPL:2015}, SMPL-X\cite{SMPL-X:2019}), which use low-dimensional parametric models as human body priors and learn to fit the model parameters to align them with the person in the picture by training. 
These template models\cite{SMPL:2015, SMPL-X:2019} derived from large amounts of data can be flexibly controlled by low-dimensional pose and shape parameters, which can capture non-rigid deformation well and reduce artifacts from linear transformation. 
The popular SMPL model learns better pose and shape blend shapes on top of linear blend skinning, and can fit various deformation through pose and shape parameters.

Most methods for reconstructing 3D humans from images are to align the distorted SMPL model with 2D images and joints by predicting the SMPL parameters \cite{2016Keep, 2020Learning}. 
However, these methods can only obtain minimally dressed human meshes, not clothed ones, due to the naked parametric model.
Other methods for reconstructing clothed human bodies extend the SMPL model into SMPL-D to represent clothes by vertex displacement\cite{bhatnagar2020ipnet, CAPE:CVPR:20}. 
IPNet\cite{bhatnagar2020ipnet} divides body and clothes into two layers, fitted with SMPL and SMPL-D respectively. 
Similarly, CAPE\cite{CAPE:CVPR:20} employs CVAE to generate corresponding meshes by pose, clothes type, and clothes shape, thereby producing the clothes vertex displacement. 
The reconstructed models can also distort the clothes mesh in different poses through the skeleton and skinning weights of the internal human model.
However, the results are often poor because wrinkles and deformation of clothes are more uncontrollable than the human body, so a lot of data is needed to learn, and corresponding scanned models are needed to supervise them. 
Clothes such as skirts and coats are also difficult to reconstruct due to the limitations of vertex displacement.

To utilize the advantages of the approaches above, recent work has attempted to combine the two representations. 
ARCH\cite{Huang_2020_arch} and ARCH++\cite{He_2021_arch++} use human prior knowledge to transform a human body in any posture into a canonical space, and then learn implicit representations for reconstruction. 
These methods produce pixel-aligned models and can theoretically be reposed by changing model parameters. 
However, since there is no learning to infer pose-dependent clothing deformation, these methods simply apply articulated deformation to the reconstructed model. 
This results in an unrealistic pose-related distortion that lacks fine details of the garment.

Since the SMPL\cite{SMPL:2015} model has only 24 joints and doesn't accommodate facial expressions and finger movements, the adoption of the SMPL-X\cite{SMPL-X:2019} model is increasingly common in the pursuit of better character fitting, which aggregates body, face, and hand. 
However, challenges arise when fitting vertices using thee SMPL-X model, including eye deformation and lip flipping. 
To address these concerns, we propose the SMPLX-Lite model, optimized for vertex fitting based on the SMPL-X, while retaining the exceptional face expression and hand action representation capabilities of the SMPL-X model.

In order to get an animatable human avatar, previous methods usually required reconstructing a character template for a single person and then modeling pose-dependent dynamic distortions. 
Recent works suggest that we can learn the deformation of a general character template from scanned data\cite{CAPE:CVPR:20, Saito:CVPR:2021SCANimate} or RGB video data\cite{liu2021neural, peng2021animatable} to get a drivable avatar directly. 
These methods usually require a large amount of data to train an avatar associated with a person, and when the data is insufficient, problems arise with over-fitting and posture generalization capabilities. 
So we introduce the SMPLX-Lite dataset, which uses 32 4K RGB cameras to capture over 20k frames of action sequences simultaneously, containing 5 characters (3 male and 2 female, wearing various types of clothes) and 15 different action types, and performs a series of data processing operations, i.e., image segmentation, 3D model reconstruction, pose estimation, SMPLX-Lite-D model fitting and texture map fitting. 
We have packaged all these annotated data into the SMPLX-Lite dataset to advance research in this field, making it possible that just a simple baseline can generate avatars with good results.

To underscore the contribution of the SMPLX-Lite dataset to the community, we develop a conditional variational autoencoder network using this dataset as a foundation following \cite{2021Driving, 2021Modeling}.  
Our method uses pose parameters, facial keypoints and view direction as conditions to generate a character model with texture based on the corresponding pose. This greatly simplifies the process of driving the character model. 
Compared with CAPE\cite{CAPE:CVPR:20}, our recovered avatar has finer geometry and photorealistic texture, making it more lifelike and directly applicable in industrial settings. 

Our contributions can be summarized as follows:
\begin{itemize}[itemsep=2pt,topsep=0pt,parsep=0pt]
    \item We collect the most comprehensive and photorealistic avatar dataset to date, containing multi-view segmented image sequences, 3D keypoint annotation, textured scanned model and fitted SMPLX-Lite-D model with texture maps. 
    
    \item We propose the SMPLX-Lite model optimized for vertex fitting based on the SMPL-X model, succeeding as the first SMPLX-based model using vertex displacement to fit clothes. 
    
    \item We introduce a multi-stage fitting procedure capturing fine geometry details like facial expressions and cloth wrinkles. 
    Compared with the SMPL-X model, it greatly reduces the difficulty of vertex fitting while retaining the details of facial expressions and hand movements.
    
    \item We propose a CVAE model that receives driving input by facial keypoints and pose parameters to produce a photorealistic avatar.
\end{itemize}

\section{SMPLX-Lite Dataset}

We present SMPLX-Lite dataset, the most comprehensive captured human avatar dataset currently. 
Please refer to the suppl. for detailed comparison with other datasets containing human model fits and a demo dataset for check.
Our dataset contains multi-view segmented image sequences, 3D keypoints annotation, reconstructed textured scanned mesh, fitted SMPLX-Lite-D model and texture maps.
In this section, we will describe in detail how to capture and organize the dataset, and the procedure for obtaining these annotation data.

\subsection{Data Capture}
\label{subsec:data capture}

We employ 32 calibrated cameras to simultaneously capture 4096x3000 image sequences of 15 different actions, being performed by 5 subjects (3 male, 2 female) in daily clothes. 
The image sequences include 15 kinds of actions in daily scenes, such as discussion, debate, public speaking, phone conversations and stretching, which significantly enhances the authentic, diverse and generalizable nature of an avatar.
For the convenience of statistics and processing, we select over 200 consecutive frames for each action sequence and eventually collect over 20k frames. 
Each frame has 32 views of the raw image, as well as all annotation results from post-processing.

\subsection{Data Process}
\label{subsec:data process}

\begin{figure}[htbp]
    \centering
    \includegraphics[width=\columnwidth]{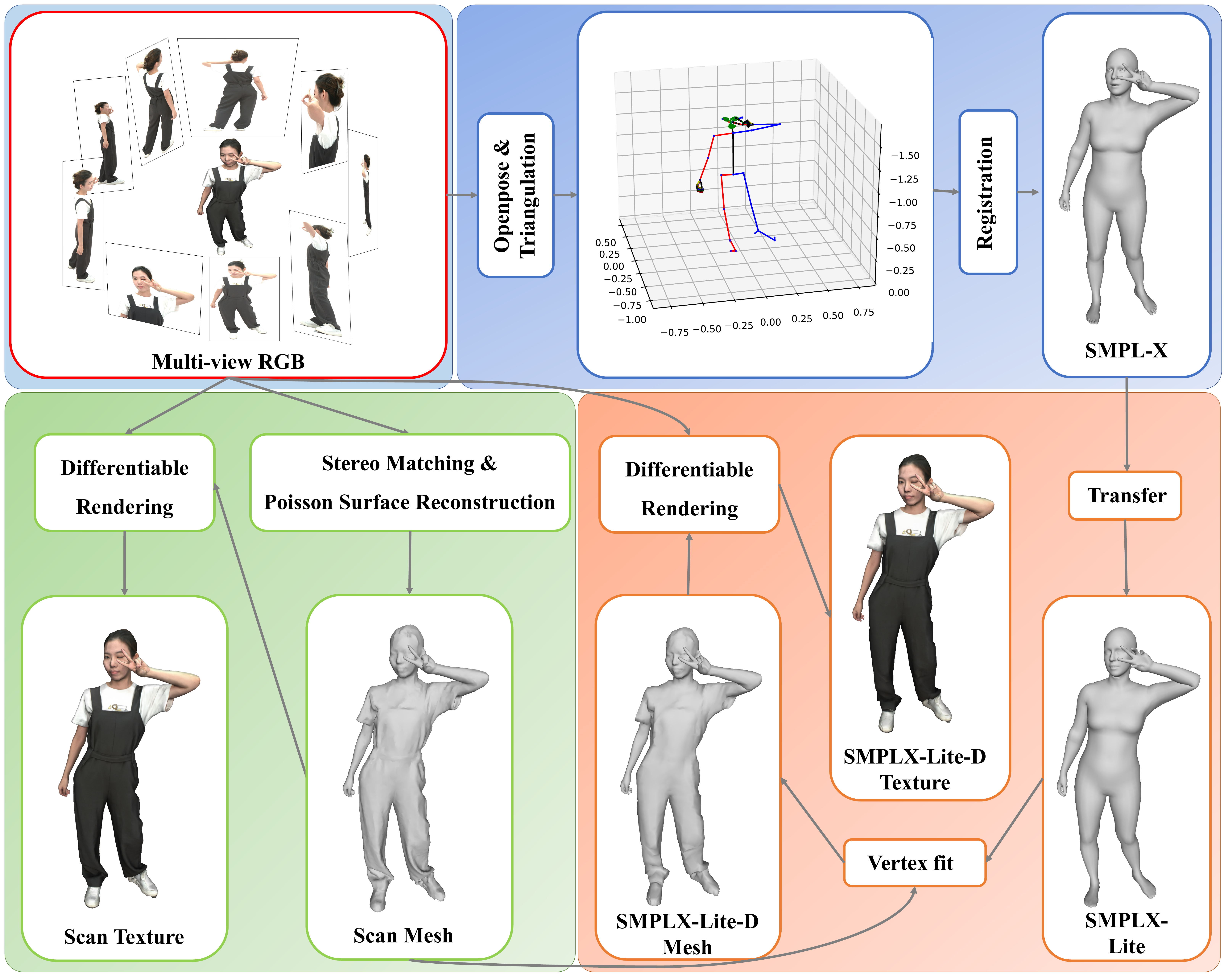}
    \caption{\textbf{Data Process Pipeline.} Our pipeline produces a variety of data annotations, including 3D keypoints, SMPL-X parameters, textured scanned models, and textured SMPLX-Lite-D models.}
    \label{fig:Data Process}
\end{figure}

\noindent\textbf{Textured Mesh Reconstruction.\label{par:scanned mesh}} \ 
We utilize 32 RGB cameras with 48 additional IR cameras and random pattern projectors for reconstruction. Following \cite{2015fvv}, we first obtain the initial depth map from the IR images through the stereo matching algorithm\cite{li2022crestereo} and then convert the depth map into a point cloud, which is later turned to the initial mesh by Poisson Surface Reconstruction (PSR) \cite{PSR2006}.
The obtained mesh has some mismatches w.r.t. the actual shape due to the accumulated error. 
We employ differentiable rendering \cite{laine2020modular} to optimize the vertex positions of the mesh geometry while extracting the texture of the mesh surface. 
Through these processes, we obtain the mesh model with higher accuracy and high-quality texture extremely close to the real picture.

\noindent\textbf{3D Human Pose Estimation.\label{par:HPE}} \ 
Once the 2D keypoints of the person from each camera view are obtained, our accurate camera intrinsic and extrinsic parameters from calibration enable the calculation of 3D keypoints by triangulation.
We use openpose\cite{openpose} to estimate 2D human joints of each view.
However, 2D keypoints estimated from different views may not be reasonable due to occlusion and limited camera field of view. 
Consequently, it is crucial to select highly confident views for each keypoint during the process of triangulation. 
We employ RANSAC\cite{1981ransac} method to select reasonable views. See suppl. for detailed process.
Subsequently, easymocap\cite{easymocap} is utilized to fit SMPL-X\cite{SMPL-X:2019} model through the supervision of 2D and 3D keypoints for every frame.

\noindent\textbf{SMPLX-Lite Model Transfer.\label{par:SMPLX-Lite transfer}} \ 
SMPL-X has $N = 10475$ vertices and $K = 54$ joints, and is defined as a function $\mathrm{M}(\theta, \beta, \psi): \mathbb{R}^{|\theta| \times|\beta| \times|\psi|} \rightarrow \mathbb{R}^{3N}$, 
where $\mathcal{\theta}$, $\mathcal{\beta}$, and $\mathcal{\psi}$ are pose, shape, and expression parameters respectively.
More specifically, 
\begin{equation}
    M(\beta, \theta, \psi)=W\left(T_p(\beta, \theta, \psi), J(\beta), \theta, \mathcal{W}\right)\label{equ:M}
\end{equation}
where $W(.)$ is a standard linear blend skinning function. Several parts of LBS function are:
\begin{equation}
    T_P(\beta, \theta, \psi)=\bar{T}+B_S(\beta ; \mathcal{S})+B_E(\psi ; \mathcal{E})+B_P(\theta ; \mathcal{P})\label{equ:T}
\end{equation}
\begin{equation}
    J(\beta)=\mathcal{J}\left(\bar{T}+B_S(\beta ; \mathcal{S})\right)\label{equ:J}
\end{equation}
and blend weights $\mathcal{W} \in \mathbb{R}^{N \times K}$. 
Methods using SMPL-X plus vertex displacement to fit clothes extend Eq.(\ref{equ:T}) to 
\begin{equation}
    T_P(\beta, \theta, \psi, D)=\bar{T}+B_S(\beta ; \mathcal{S})+B_E(\psi ; \mathcal{E})+B_P(\theta ; \mathcal{P})+D.\label{equ:T2}
\end{equation}

\begin{figure}[tbp]
  \centering
  \mbox{}\hfill
  \subfloat[SMPL-X vertex fit.]{\label{fig:smplx fit} \includegraphics[width=0.21\textwidth]{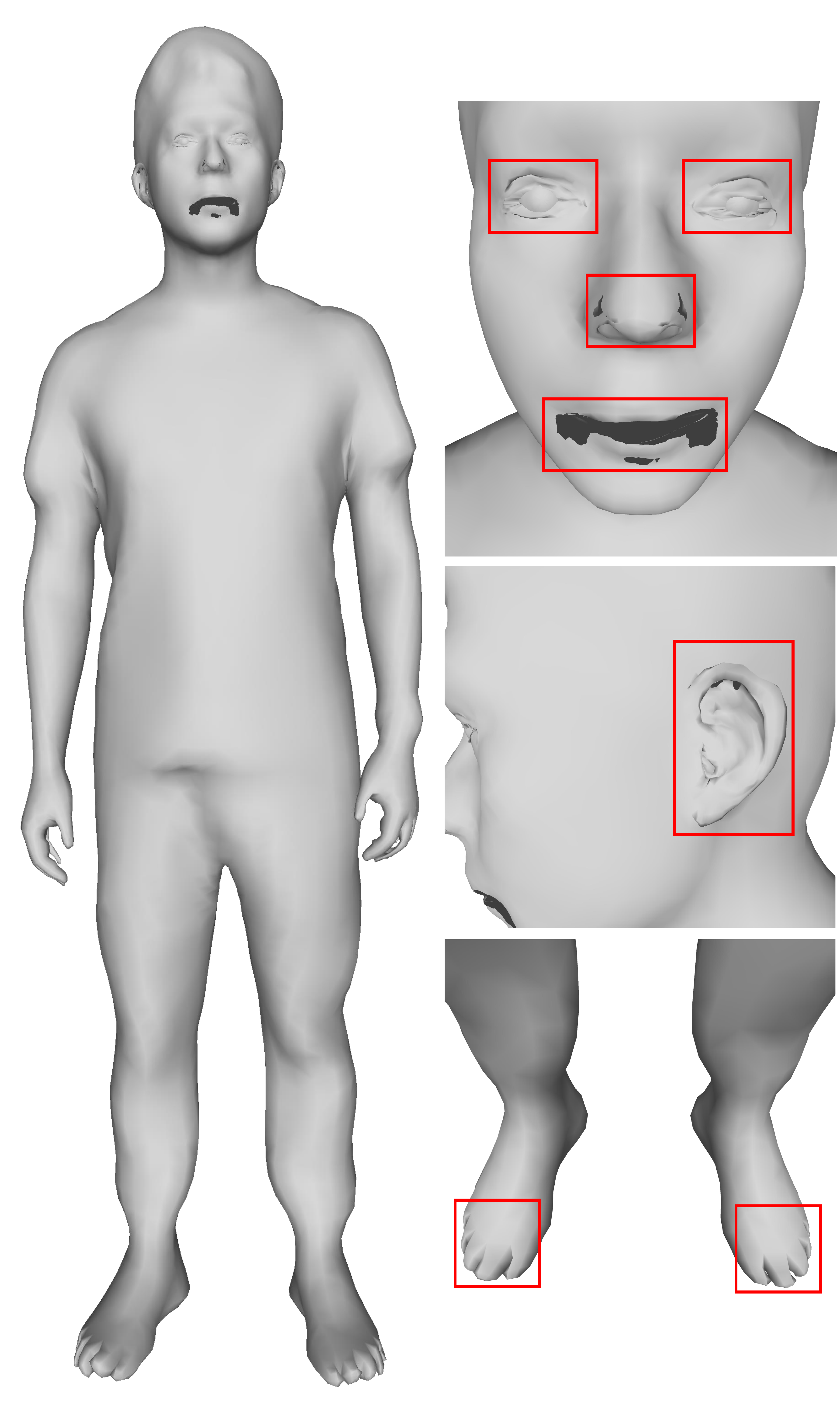}}
  \hfill
  \subfloat[SMPLX-Lite vertex fit.]{\label{fig:smplx-lite fit} \includegraphics[width=0.21\textwidth]{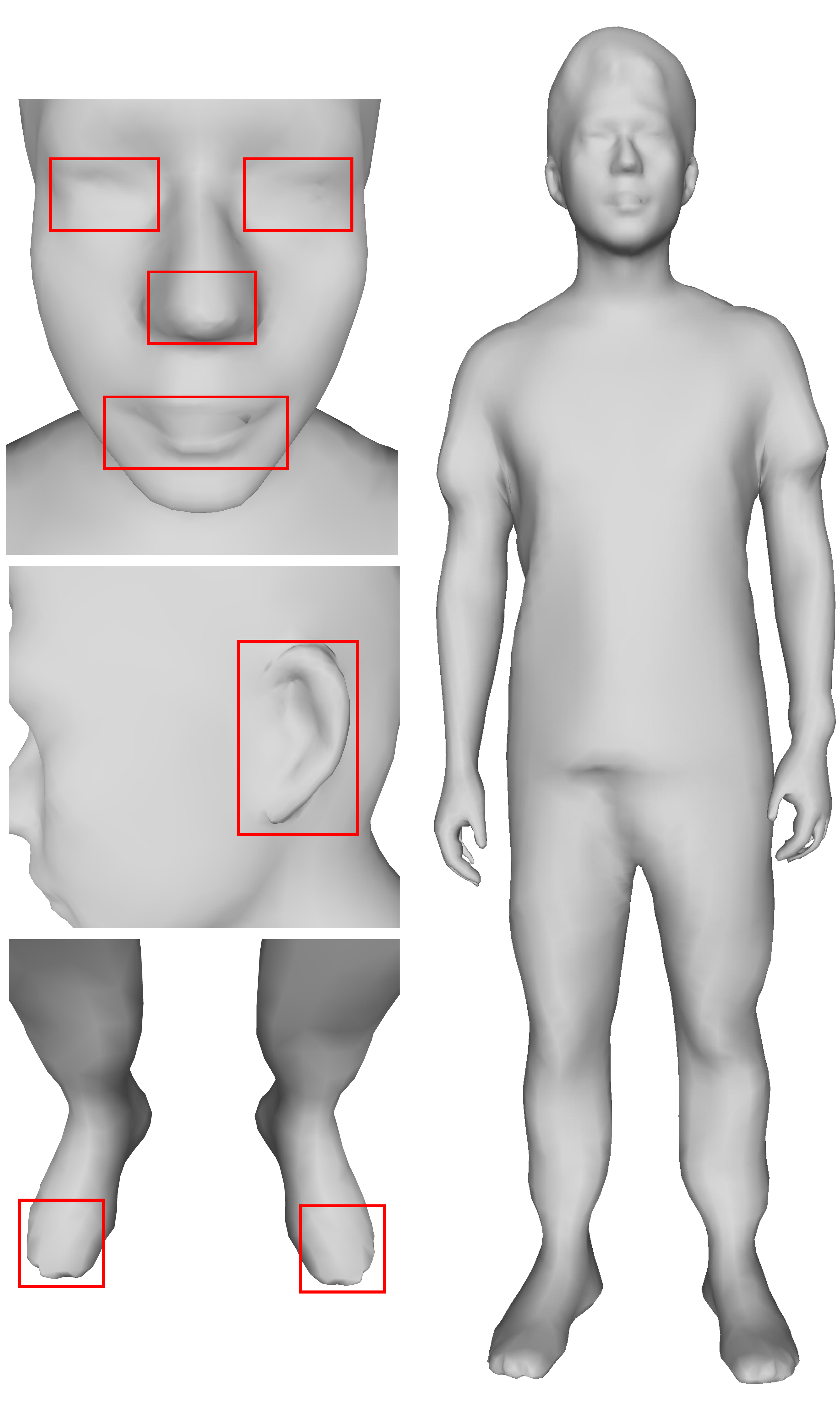}}
  \hfill\mbox{}
  \caption{Comparison between SMPL-X and SMPLX-Lite model fit results.}
  \label{fig:fit comparison}
\end{figure}

The SMPL-X model with vertex displacement shown in Fig.\ref{fig:smplx fit}, exhibits face flipping and distortion in the eyes, ears, mouth, nose and feet. 
In response to these issues, we propose SMPLX-Lite model, which greatly reduces the difficulty of vertex fitting while preserving the facial expression and hand gesture fitting capabilities of the SMPL-X model. 
The iterative process entail vertex deletion, face reconstruction, and face flattening, ultimately yielding the SMPLX-Lite model with a reduced vertex count of $N_{v} = 8452$.
Refer to suppl. for details.

As the number of vertices decreases, adjustments to the matrices $\mathcal{S}$, $\mathcal{E}$, $\mathcal{P}$, $\mathcal{J}$, $\mathcal{W}$, as indicated in Eq.(\ref{equ:M}, \ref{equ:T}, \ref{equ:J}), are vital for ensuring the transferred model inherits the control parameters of SMPL-X and the linear blend skinning function.
Upon transferring all coefficient matrices, the SMPLX-Lite model becomes operational akin to SMPL-X, and utilizing Eq.(\ref{equ:T2}), vertex displacement can be added to the model to fit clothes.
The subsequent analysis will demonstrate the impressive efficacy of this model in vertex fitting.

\noindent\textbf{SMPLX-Lite-D fit.\label{par:SMPLX-Lite-D fit}}\ 
The purpose of vertex fit is to fully capture the fine geometry details of the scanned meshes in a unified mesh topology and texture UV layout. 
After the 3D pose estimation in section\ref{par:HPE}, we obtain a starting mesh close to the scanned mesh without surface details. 
We propose to solve for vertex fit in \textbf{2} stages. 
In the first stage, we adopt the method from \cite{2009Robust} and warp the mesh by predefined embedded nodes, then solve for the warp field. 
In the second stage, we directly solve for the remaining vertex shifts.
The detailed procedures and impact of all the registration steps are illustrated in suppl.

\noindent\textbf{How will this dataset be useful to the community?} \ 
Dedicated significant effort has been made to collect and process the most comprehensive 3D moving human avatar dataset with clothes and textures to date. 
The SMPLX-Lite dataset has significant implications for \textbf{Drivable Textured Avatar Reconstruction}, as it provides multi-view images, reconstructed texture models, and fitted clothed parametric models with texture maps. 
These diverse data types can be leveraged to reconstruct photorealistic drivable avatars, offering researchers a wider spectrum of supervising methods compared to datasets that offer only raw pictures \cite{cheng2022generalizable} or solely reconstructed textured models \cite{CAPE:CVPR:20}. 
This capability broadens the range of network structures that can be utilized, potentially enabling multiple stages of network training. 

Besides, the SMPLX-Lite dataset is also pertinent to other important areas such as \textbf{3D Human Body Reconstruction} and  \textbf{Novel View Synthesis}.
Moreover, researchers are encouraged to explore further applications of this dataset.

\subsection{Dataset Evaluation}
\label{subsec:dataset eval}

\begin{table}[tbp]
    \caption{\textbf{Dataset Evaluation Results.} We render textured models, compare them with captured images, and compare the geometry of the fitted SMPLX-Lite-D model with scanned mesh to get chamfer distance (CD, $\times 10^{-3}$).}
	\centering
	\resizebox{0.45\textwidth}{!}{
	\begin{tabular}{lccccc}
		\toprule
		        &    \multicolumn{2}{c}{Scan}     &    \multicolumn{2}{c}{SMPLX-Lite-D}     &                                  \\
		\cmidrule(r){2-3} \cmidrule(r){4-5}
		Sub. & PSNR$\uparrow$ & SSIM$\uparrow$ & PSNR$\uparrow$ & SSIM$\uparrow$ & CD$\downarrow$ \\
		\midrule
		WZL     &     28.92      &     0.9714     &    \textbf{28.61}      &     0.9706     &              6.7372              \\
		LDF     &     28.33      &     0.9706     &     27.80      &     0.9675     &              8.2897              \\
		ZX      &     \textbf{28.95}      &     \textbf{0.9760}     &     28.52      &     \textbf{0.9749}     &              6.8234              \\
		LW      &     27.21      &     0.9754     &     26.67      &     0.9744     &              \textbf{6.4386}              \\
		ZC      &     27.51      &     0.9623     &     27.05      &     0.9602     &              6.9238              \\
		\bottomrule
	\end{tabular}}

	\label{tab:dataset eval}
\end{table}

We present the evaluation results in Tab.\ref{tab:dataset eval}, including peak signal-to-noise ratio (PSNR), structural similarity index measure (SSIM), and chamfer distance (CD). 

\section{Method}
To demonstrate the effectiveness of SMPLX-Lite dataset, we utilize a basic baseline model to generate a drivable avatar and show the effect that all avatars in the dataset are driven by the same sequence of actions.
Our approach is grounded on the network structure in \cite{2021Driving, 2021Modeling}, involving comprehensive simplifications and adaptations. 

\subsection{Network Structure}
\label{subsec:network}

\begin{figure}[htbp]
  \centering
    \includegraphics[width=\columnwidth]{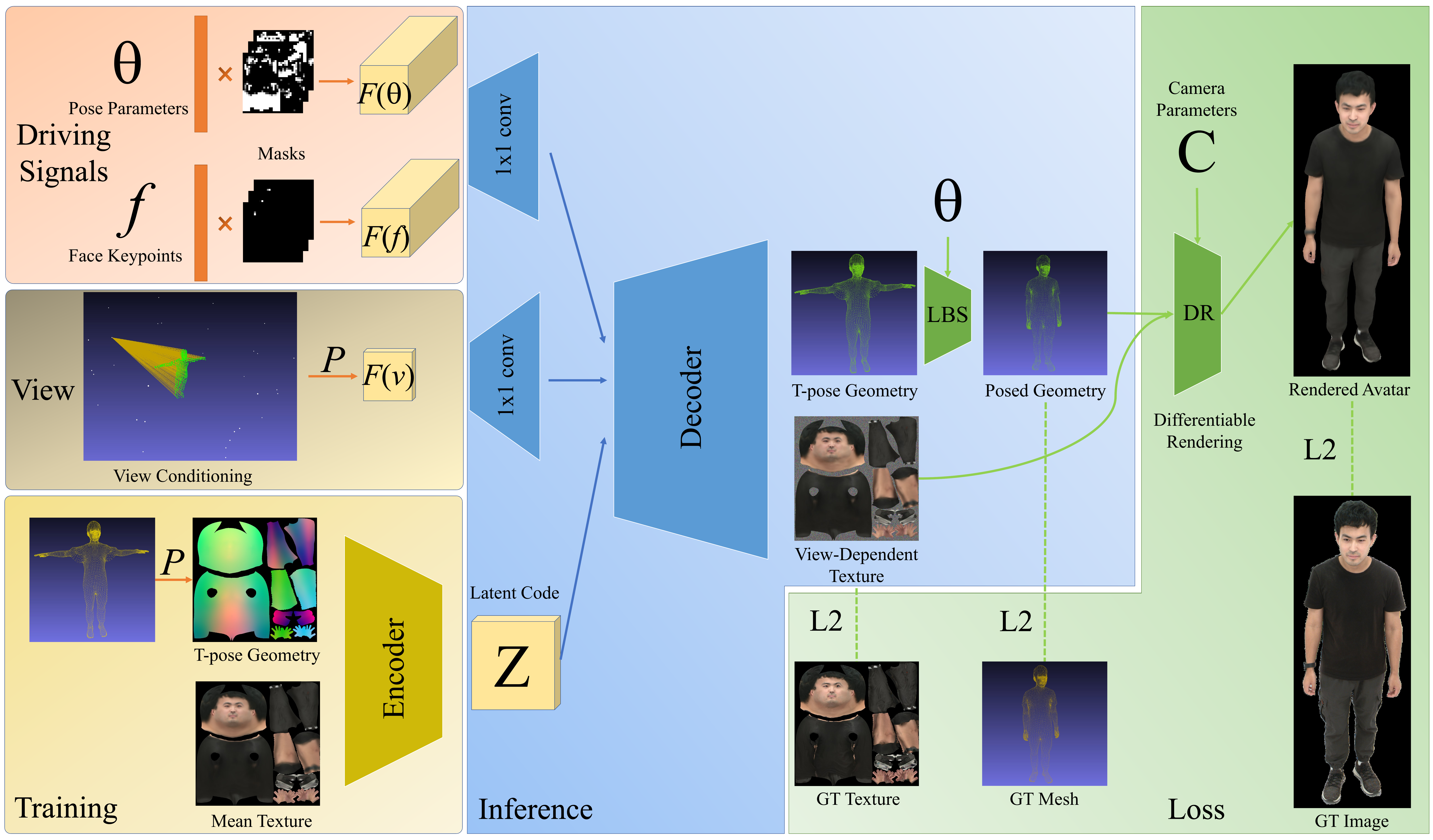}
    \caption{\textbf{Method Overview.} The CVAE model generates mesh and texture maps via a decoder, which employs pose and face keypoints as driving signals, overlays camera view information, and utilizes latent codes sampled from the distribution obtained by the encoder. The output mesh obtained by LBS, together with the texture map and camera parameters, undergoes the differentiable renderer to produce photorealistic rendered images. The entire training process is end-to-end, and mesh, texture, and final rendered images are all supervisable.}
    \label{fig:method}
\end{figure}

The model employed is a conditional variational autoencoder (CVAE), consisting of an encoder $E$ and a decoder $D$, both implemented using convolutional neural networks.
See Fig.\ref{fig:method} for the overview of our method. 

The encoder takes as input the mean texture map $\bar{\boldsymbol{T}}$ for each individual in the dataset and the T-pose mesh $\boldsymbol{G_{i}}\in\mathbb{R}^{N\times3}$ derived via inverse LBS to SMPLX-Lite-D model for each frame. 
Rendering $\boldsymbol{G_{i}}$ onto a position map in UV space yields a feature map of the same size as the average texture map, which is subsequently merged with the average texture map in channel dimension and fed into the encoder. 
$$
E(\boldsymbol{T}, P(\boldsymbol{G_{i}})) \rightarrow \boldsymbol{\mu_{i}}, \boldsymbol{\sigma_{i}}
$$
The encoder outputs the mean $\boldsymbol{\mu_{i}}$ and standard deviation $\boldsymbol{\sigma_{i}}$ of the Gaussian distribution, which are trained to align as closely as possible to the standard normal distribution $\mathcal{N}(0, 1)$ and then sampled to obtain the latent code $\boldsymbol{z}$.
$$
\boldsymbol{z_{i}} \sim \boldsymbol{\mathcal{N}}(\boldsymbol{\mu_{i}}, \boldsymbol{\sigma_{i}}^{2})
$$
Following \cite{2021Driving}, we utilize readily available pose parameters $\boldsymbol{\theta_{i}}$ and face keypoints $\boldsymbol{f_{i}}$ as driving signals. 
These are rendered to a position map in UV space and merged into feature maps, while the T-pose vertex coordinate is leveraged to generate view information feature maps. 
These driving signals and view feature maps serve as conditions and are combined with the latent code before being fed into the decoder to predict the T-pose mesh offsets $\boldsymbol{\hat{G}^{T}_{i}}$ and view-dependent texture map $\boldsymbol{\hat{T}_{i}}$:
$$
D(W(\boldsymbol{\mathcal{F}_{\theta_{i}}}, \boldsymbol{\mathcal{F}_{f_{i}}}, \boldsymbol{\mathcal{F}_{v_{i}}}), \boldsymbol{z_{i}}) \rightarrow \boldsymbol{\hat{G}^{T}_{i}}, \boldsymbol{\hat{T}_{i}}, 
$$
where $W$ means conv1x1.
We use decoder to predict offsets because the network fits the residuals better than directly fitting the vertex locations. 
By adding these offsets to T-pose template $\boldsymbol{\bar{G}_{T}}$ and transforming them using pose parameters $\boldsymbol{\theta_{i}}$ through LBS, the final reconstructed pose mesh $\boldsymbol{\hat{G}_{i}}$ is obtained:
\begin{equation}
    \boldsymbol{\hat{G}_{i}} = LBS(\boldsymbol{\theta}, \boldsymbol{\bar{G}_{T}} + \boldsymbol{\hat{G}^{T}_{i}}). 
    \label{equ:LBS}
\end{equation}

The dataset provided allows for the supervision of $\boldsymbol{\hat{G}_{i}}$ and $\boldsymbol{\hat{T}_{i}}$ through the model geometry of SMPLX-Lite-D and the associated texture map, as well as the rendered images $\boldsymbol{\hat{I}_{i}}$ through differentiable rendering with the captured images $\boldsymbol{I_{i}}$. 
This multi-faceted supervision facilitates the creation of high-quality drivable reconstructed human models.

During inference, the latent code $\boldsymbol{z}$ is sampled from a standard normal distribution without the need for an encoder.
Decoder $D$ takes $\boldsymbol{z}$ along with the driving signal and view information as input to generate the geometry and texture of the person under the corresponding pose.

\subsection{Loss}
\label{subsec:loss}

The loss function we use is:
\begin{equation}
    \mathcal{L}=\lambda_{G}\mathcal{L}_{G} + \lambda_{T}\mathcal{L}_{T} + \lambda_{lap}\mathcal{L}_{lap} + \lambda_{KL}\mathcal{L}_{KL}, \label{equ:Loss}
\end{equation}
where $\mathcal{L}_{\boldsymbol{G}} = ||\boldsymbol{G}_{gt} - \hat{\boldsymbol{G}}||_{2}^{2}$ is the L2 distance between the vertex of the reconstructed model and gt SMPLX-Lite-D model,
$\mathcal{L}_{\boldsymbol{T}} = ||(\boldsymbol{T}_{gt} - \hat{\boldsymbol{T}}) \odot \boldsymbol{M_{T}}||_{2}^{2}$ is the L2 loss of the texture map and gt texture map in the valid UV area with mask $M_{T}$, 
$\mathcal{L}_{lap} = ||L(\boldsymbol{G}_{gt}) - L(\hat{\boldsymbol{G}})||_{2}^{2}$ is the Laplacian term used to ensure the smoothness of the model,
and $\mathcal{L}_{KL}$ is the KL term of the standard VAE model\cite{kingma2013autoencoding}.
If gt image is used for supervision, $\mathcal{L}_{T}$ can be replaced with image loss $\mathcal{L}_{\boldsymbol{I}} = ||\boldsymbol{I}_{gt} - \hat{\boldsymbol{I}}||_{2}^{2}$ plus image mask loss $\mathcal{L}_{\boldsymbol{M}} = ||\boldsymbol{M}_{gt} - \hat{\boldsymbol{M_{I}}}||_{2}^{2}$.

\section{Experiments}
\label{sec:experiments}
In this section, we present the results of our photorealistic human model-driven algorithm on the SMPLX-Lite dataset. 
Subsequently, we compare our method with two baselines, in both novel view and novel pose synthesis experiments, to demonstrate the superior performance of our method in geometry and texture generation.

\subsection{Reconstruction \& Driving}
\label{subsec:reconstruction}

We utilize 9 actions in the dataset for training and the others for testing. 
Subsequent experiments involve training individual character models on the training set  and evaluating their reconstruction and driving effects on the test set. 

To begin, we assess the method's ability to reconstruct mesh and texture for new actions of the same person on the test set, which involves utilizing the encoder $E$ to generate latent code $\boldsymbol{z}$ with the same distribution as the training data. 
Additionally, we evaluate the driving effect of the model by using the driving signal of the test set to drive the characters. 
Unlike reconstruction, driving necessitates the random sampling of latent code $\boldsymbol{z}$ from the normal distribution without encoder $E$. 

\begin{figure}[htbp]
    \centering
    \mbox{}\hfill
    \subfloat[Recon]{\label{fig:Reconstruction} \includegraphics[width=0.08\textwidth]{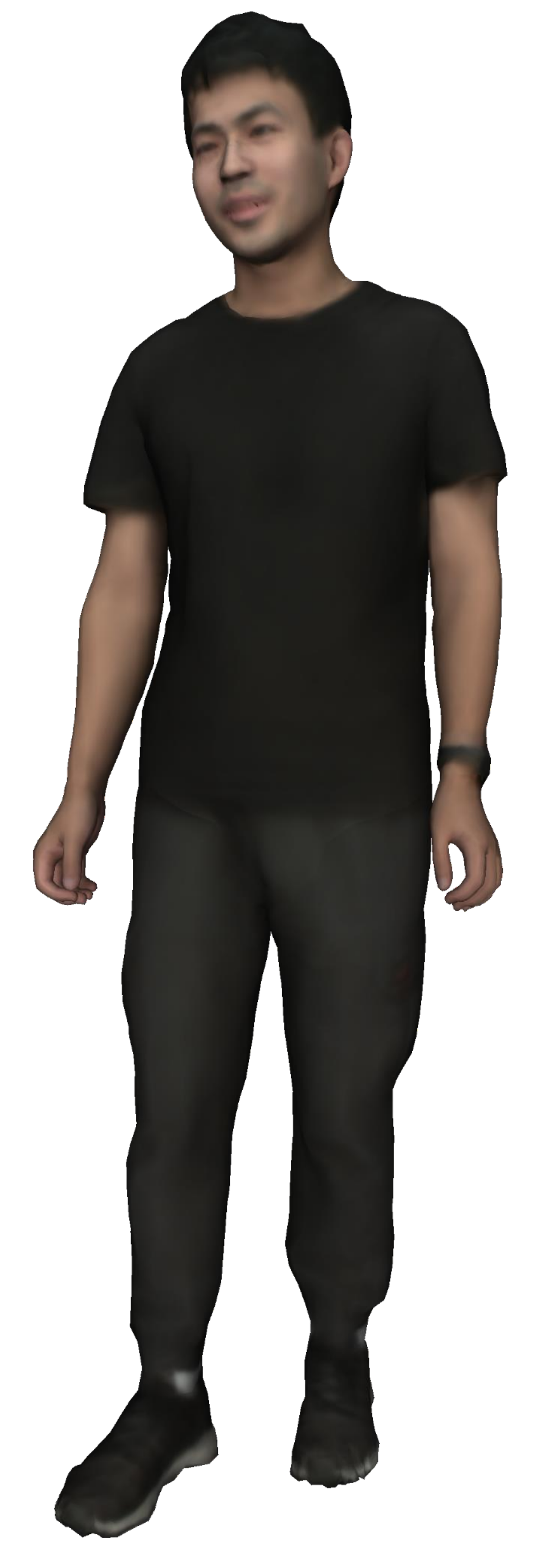}}
    \hfill
    \subfloat[Driving]{\label{fig:Driving} \includegraphics[width=0.08\textwidth]{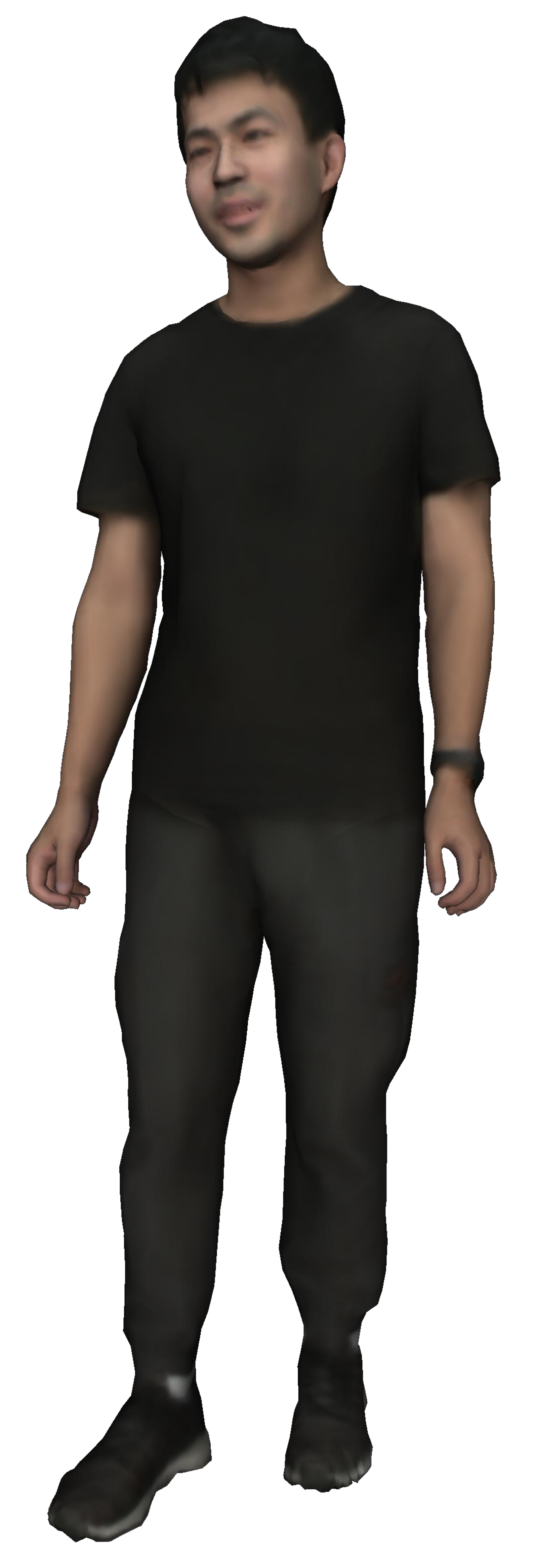}}
    \hfill
    \subfloat[GT]{\label{fig:GT Image} \includegraphics[width=0.08\textwidth]{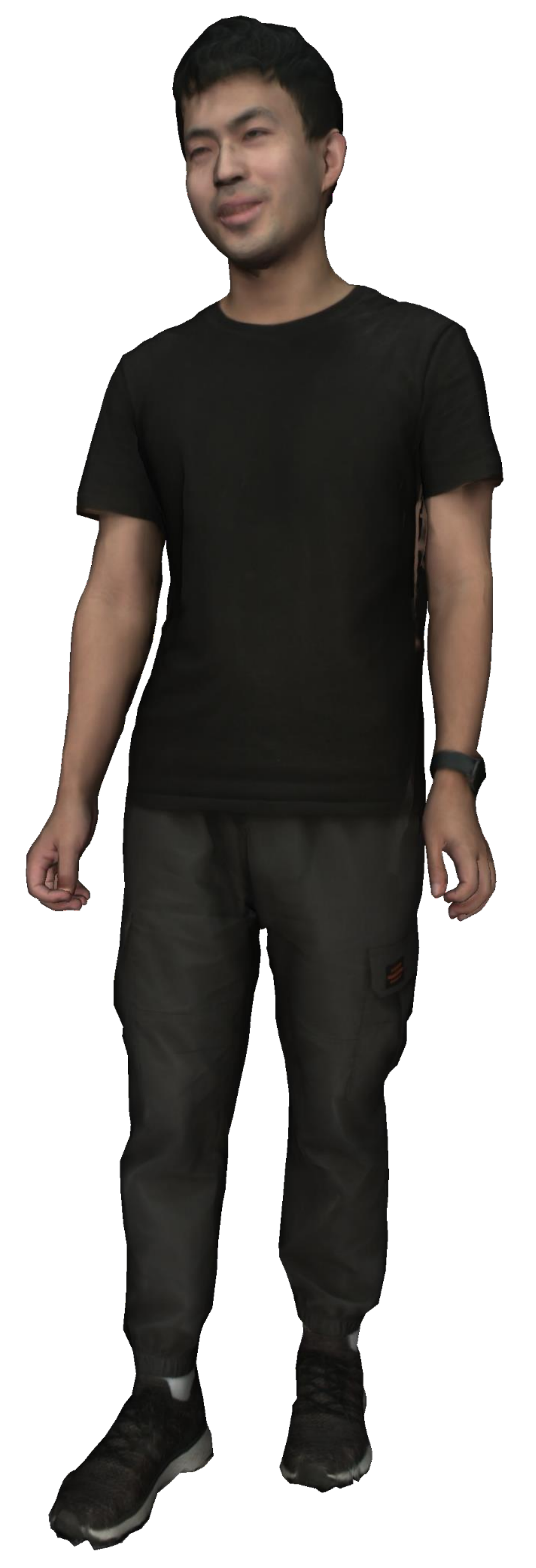}}
    \hfill\mbox{}

    \caption{\textbf{Qualitative Results.} Both rendered images are really close to the captured image, perfectly recovering clothing details, finger movements and facial expressions.}
    \label{fig:driving results}
\end{figure}

The photorealistic reconstructed and driving results, along with quantitative evaluations for all subjects, are presented in Fig.\ref{fig:driving results} and Tab.\ref{tab:driving eval}, respectively.
It is worth noting that driving is marginally less effective than reconstructing due to the absence of hidden space information associated with the character.
Furthermore, we test the effect of using the same new sequence of actions to drive five trained character models and present the full results in suppl.

\begin{table}[htbp]
    \caption{\textbf{Quantitative Evaluation.} CD ($\times 10^{-3}$) is the distance between the generated model and the SMPLX-Lite-D model.}
	\centering
	\resizebox{0.45\textwidth}{!}{
		\begin{tabular}{lcccccc}
			\toprule
			                                   &    \multicolumn{2}{c}{Reconstruction}     && \multicolumn{2}{c}{Driving} &                \\
			\cmidrule
			(r){2-3} \cmidrule(r){5-6}
			Sub. & PSNR$\uparrow$ & SSIM$\uparrow$ & CD$\downarrow$&PSNR$\uparrow$ & SSIM$\uparrow$  & CD$\downarrow$ \\
			\midrule
			WZL                                &     \textbf{26.60}      &     \textbf{0.9454}     &  4.1098&   \textbf{26.54}      &     \textbf{0.9443}      &     \textbf{4.2264}     \\
			LDF                                &     25.69      &     0.9394     & \textbf{3.9842}&    24.94      &     0.9307      &    4.2419     \\
			ZX                                 &     25.33      &     0.9382     &  4.7422&   24.71      &     0.9312      &     4.7719     \\
			LW                                 &     23.38      &     0.9397     & 5.2762&    22.42      &     0.9304      &     5.4222     \\
			ZC                                 &     24.48      &     0.935     &  4.6135&   23.49      &     0.9254      &     4.6383     \\
			\bottomrule
		\end{tabular}}

	\label{tab:driving eval}
\end{table}

\subsection{Comparison with Baselines}

\begin{figure}[htbp]
    \centering
    \subfloat[Novel View Synthesis]{\label{fig:Novel View} \includegraphics[width=0.48\textwidth]{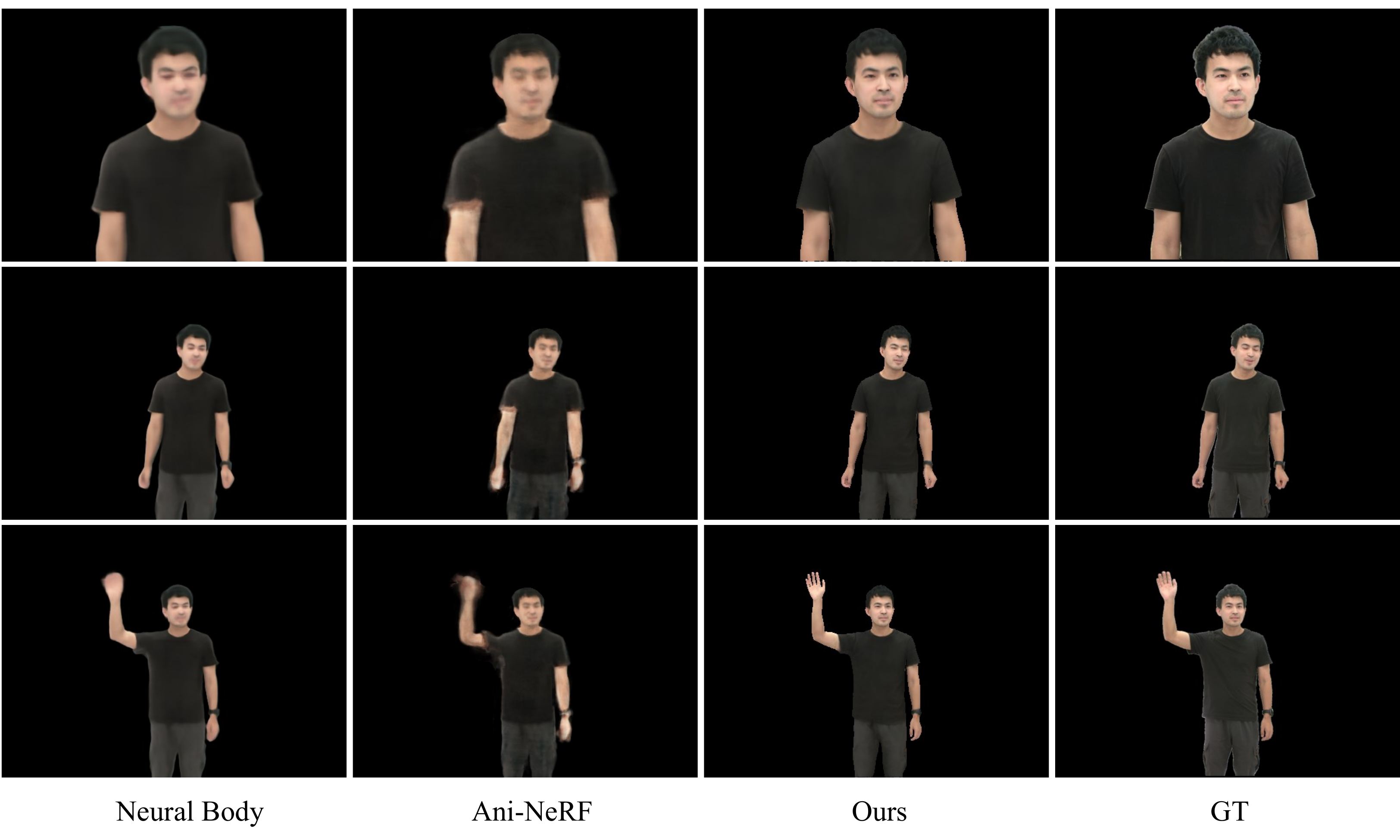}}
    \vspace{0pt}
    \subfloat[Novel Pose Synthesis]{\label{fig:Novel Pose} \includegraphics[width=0.48\textwidth]{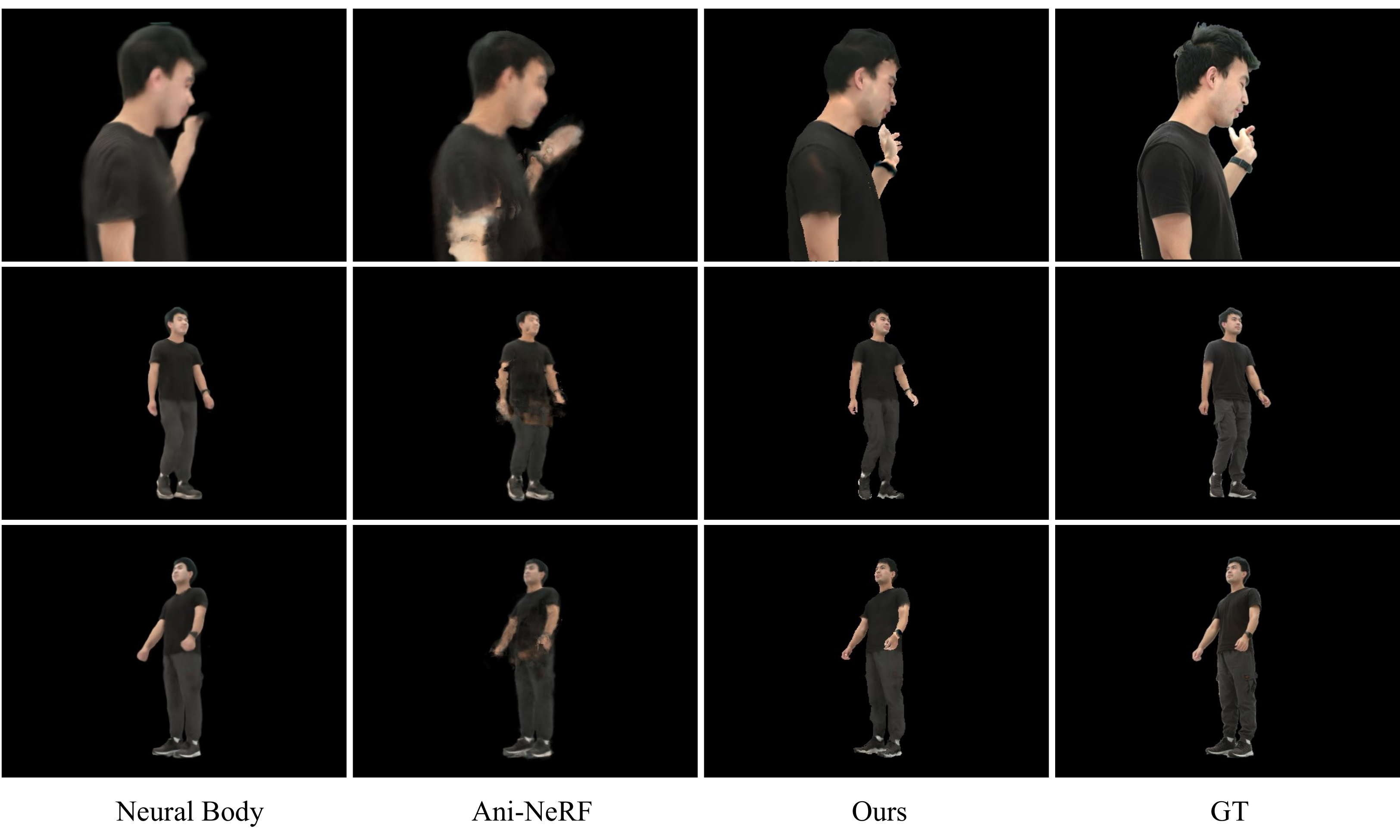}}

    \caption{Our results in both experiments show clearer and more realistic textures and accurately reconstruct finger movements.}
    \label{fig:Novel results}
\vspace{-1.0em}
\end{figure}

Additionally, We conduct comparisons with two baselines, Neural Body (NB)\cite{peng2021neural} and Ani-NeRF (AN)\cite{peng2021animatable}. 
Following NB's setting, our method outperforms the two baselines in both novel view and novel pose synthesis experiments, as demonstrated in Fig.\ref{fig:Novel results} and Tab.\ref{tab:results}.

\begin{table}[htbp]
    \caption{\textbf{Quantitative Results.} Our method outperforms the baselines in terms of  LPIPS and chamfer distance(CD, $\times 10^{-3}$), while also achieving competitive PSNR and SSIM.}
	\centering
     \vspace{4pt}
	\resizebox{0.48\textwidth}{!}{
	\begin{tabular}{lcccccccc}
    \hline
    \multicolumn{1}{c}{\multirow{2}{*}{Method}} & \multicolumn{4}{c}{Novel View}                                         & \multicolumn{4}{c}{Novel Pose}                                         \\ \cline{2-9} 
    \multicolumn{1}{c}{}                        & PSNR$\uparrow$ & SSIM$\uparrow$  & LPIPS$\downarrow$ & CD$\downarrow$  & PSNR$\uparrow$ & SSIM$\uparrow$  & LPIPS$\downarrow$ & CD$\downarrow$  \\ \hline
    NB                                          & \textbf{31.29} & \textbf{0.9707} & {\ul 0.0789}      & {\ul 11.490}    & \textbf{29.27} & \textbf{0.9616} & {\ul 0.0841}      & {\ul 11.732}    \\
    AN                                          & 28.05          & 0.9500          & 0.0981            & 16.285          & 26.14          & 0.9382          & 0.1119            & 18.170          \\
    Ours                                        & {\ul 30.14}    & {\ul 0.9607}    & \textbf{0.0567}   & \textbf{7.1586} & {\ul 27.97}    & {\ul 0.9485}    & \textbf{0.0675}   & \textbf{8.7690} \\ \hline
    \end{tabular}
    }
	\label{tab:results}
\end{table}

\begin{figure}[htbp]
    \centering
    \includegraphics[width=0.8\linewidth]{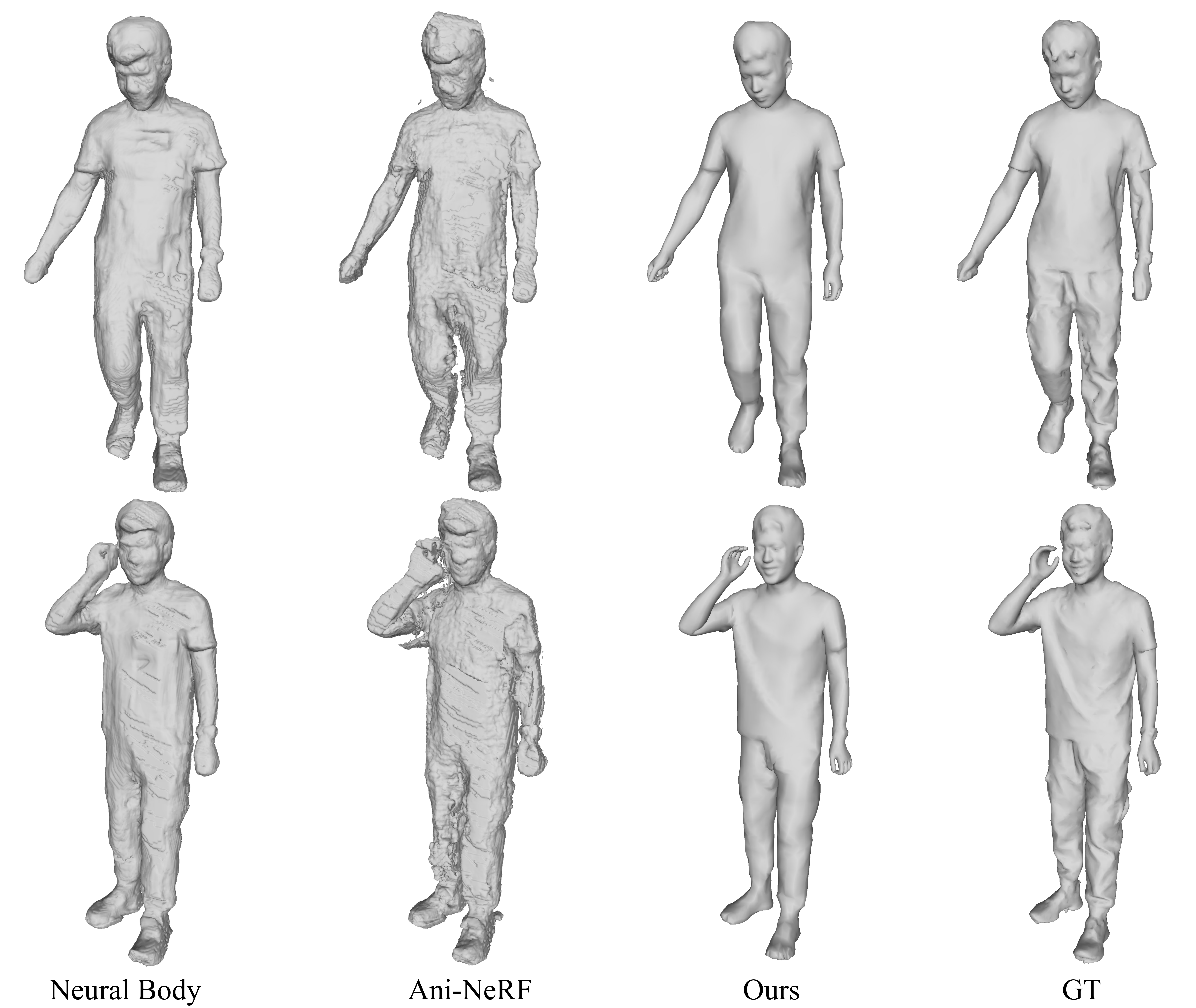}
    \caption{\textbf{Mesh Generation}. Our method enables the reconstruction of smoother surface and finer geometric details.}
    \label{fig:Mesh}
\vspace{-1.0em}
\end{figure}

The robust and highly generalizable nature of our approach enables it to capture intricate details and high-frequency information, leading to clearer textures and hand movements. In contrast, the baselines produce notably blurry results in both experiments, particularly in the hand area, with AN displaying abnormally twisted arms and fingers in Fig.\ref{fig:Novel Pose}. 
Besides, the meshes generated by our method appear smoother and retain a higher level of detail, as depicted in Fig.\ref{fig:Mesh}. %


\section{Conclusion}
\label{sec:conclusion}
We propose the SMPLX-Lite model, which simplifies the methods using vertex displacement to fit clothes, while retaining the advantages of the SMPL-X model. 
This paves the way for the generation of the proposed SMPLX-Lite dataset, which stands as the most comprehensive and fairly photorealistic textured clothed avatar dataset currently available, supporting the advancement of the research community. 
Leveraging this dataset, we introduce a CVAE-based textured human model driving algorithm, showcasing the substantial advantage of SMPLX-Lite dataset in label richness and photorealism.
Notably, our driving algorithm utilizes solely the captured images and textured SMPLX-Lite-D model in the dataset. 
Additionally, the SMPLX-Lite dataset includes annotations for 2D/3D keypoints and SMPL-X model, high-precision scanned models, and corresponding texture maps, which are invaluable data contributing to pertinent research endeavors.

\section*{Acknowledgment}
This work was partly supported by the National Natural Science Foundation of China under
U23B2030 and the Special Foundations for the Development of Strategic Emerging Industries of Shenzhen (Nos.JSGG20211108092812020 \& CJGJZD20210408092804011).

\bibliographystyle{IEEEtran}
\bibliography{IEEEexample} 

\clearpage
\appendices

\begin{figure*}[ht]
    \centering
    \includegraphics[width=0.95\textwidth]{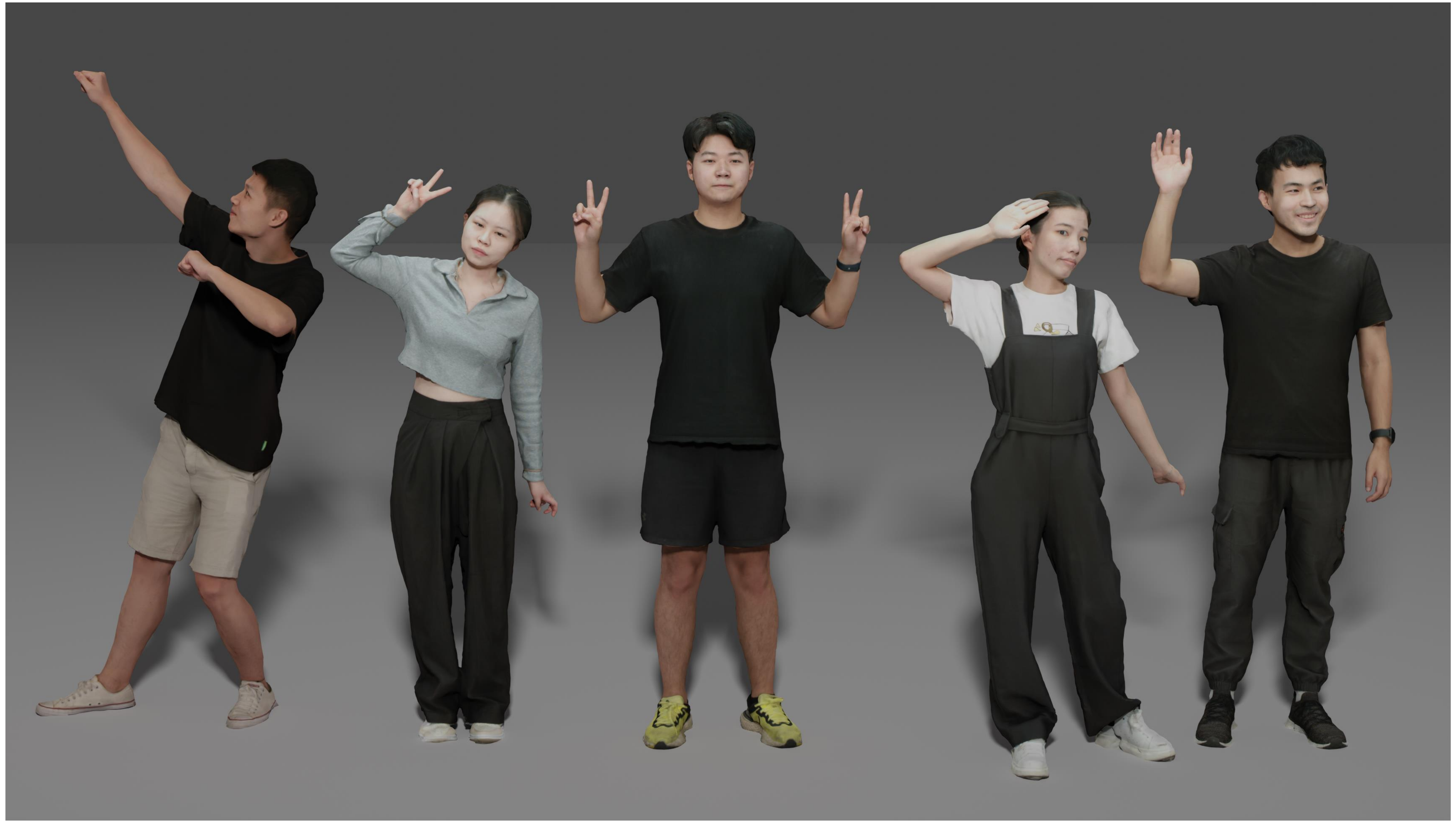}
    \caption{SMPLX-Lite provides over 20k high-resolution scan models of 5 subjects performing 15 types of actions.}
    \label{fig:all}
\end{figure*}

We provide additional dataset processing details (Appendix~\ref{sec:dataset}), extended dataset evaluation results (Appendix~\ref{sec:eval}), extended diverse dataset visualization (Appendix~\ref{sec:vis}) and extended experiment results (Appendix~\ref{sec:experiments_sup}).  
Actors and actresses participating in SMPLX-Lite are well-informed and acknowledge that the data will be made public for research purposes.
%
\section{Additional Dataset Processing Details}
\label{sec:dataset}

\subsection{3D Human Pose Estimation}
\label{sec:HPE}
We provide a detailed description of the RANSAC\cite{1981ransac} algorithm mentioned in Sec.3.2 of the main paper in Algorithm\ref{alg1}.

\begin{algorithm}\small
    \renewcommand{\algorithmicrequire}{\textbf{Input:}}
    \renewcommand{\algorithmicensure}{\textbf{Output:}}
    \caption{3D Keypoints Estimation by RANSAC} 
    \label{alg1}
    \begin{algorithmic}[1]
        \REQUIRE Detected 2D Keypoints $K_{2D}$, camera views $C$, number of keypoints $J$, projection matrix $P$, reprojection threshold $\tau$, RANSAC confidence $p$, sample views $V$, number of sample views $v$, maximum number of iterations $I$, minimum reprojection error $e$, iteration $i$
        \ENSURE Estimated 3D Keypoints $K_{3D}$, reasonable 2D reprojection keypoints $K_{repro}$
        
        \FORALL{$j \in range(0, J)$}
            \STATE $I = 10000, i = 0, e = 1000$
            \WHILE{$i \leq I$} 
                \STATE $V = RANDOM\_SELECT(C, v)$
                \STATE $\hat{K_{3D}^{j}} = TRIANGULATE(K_{2D, V}^{j}, P_{V})$
                \STATE $\hat{K_{repro}^{j}} = REPROJECTION(\hat{K_{3D}^{j}}, P)$
                \STATE $D_{V} = DISTANCE(K_{2D}^{j}, \hat{K_{repro}^{j}})$
                \STATE $\hat{C}, N_{\hat{C}} = SELECT(D, \tau)$
                \IF{$N_{\hat{C}} > 3$} 
                    \STATE $\hat{K_{3D}^{j}} = TRIANGULATE(K_{2D, \hat{C}}^{j}, P_{\hat{C}})$
                    \STATE $\hat{K_{repro}^{j}} = REPROJECTION(\hat{K_{3D}^{j}}, P)$
                    \STATE $D = MEAN\_DISTANCE(K_{2D}^{j}, \hat{K_{repro}^{j}})$
                    \IF{$D < e$}
                        \STATE $e = D, Inliers = N_{\hat{C}}$ 
                        \STATE $I = \frac{log(1 - p)}{log(1 - (Inliers / |C|)^{v}}$
                        \STATE $K_{3D}^{j} = \hat{K_{3D}^{j}}, K_{repro}^{j} = \hat{K_{repro}^{j}}$
                    \ENDIF
                \ENDIF 
                \STATE i += 1
            \ENDWHILE
        \ENDFOR
        \RETURN $K_{3D}$, $K_{repro}$
    \end{algorithmic}
\end{algorithm}

\subsection{SMPLX-Lite Model Transfer}
\label{sec:transfer}

\begin{figure*}[ht]
    \centering
      
      \subfloat[SMPL-X]{\label{fig:transfer1} \includegraphics[width=0.32\textwidth]{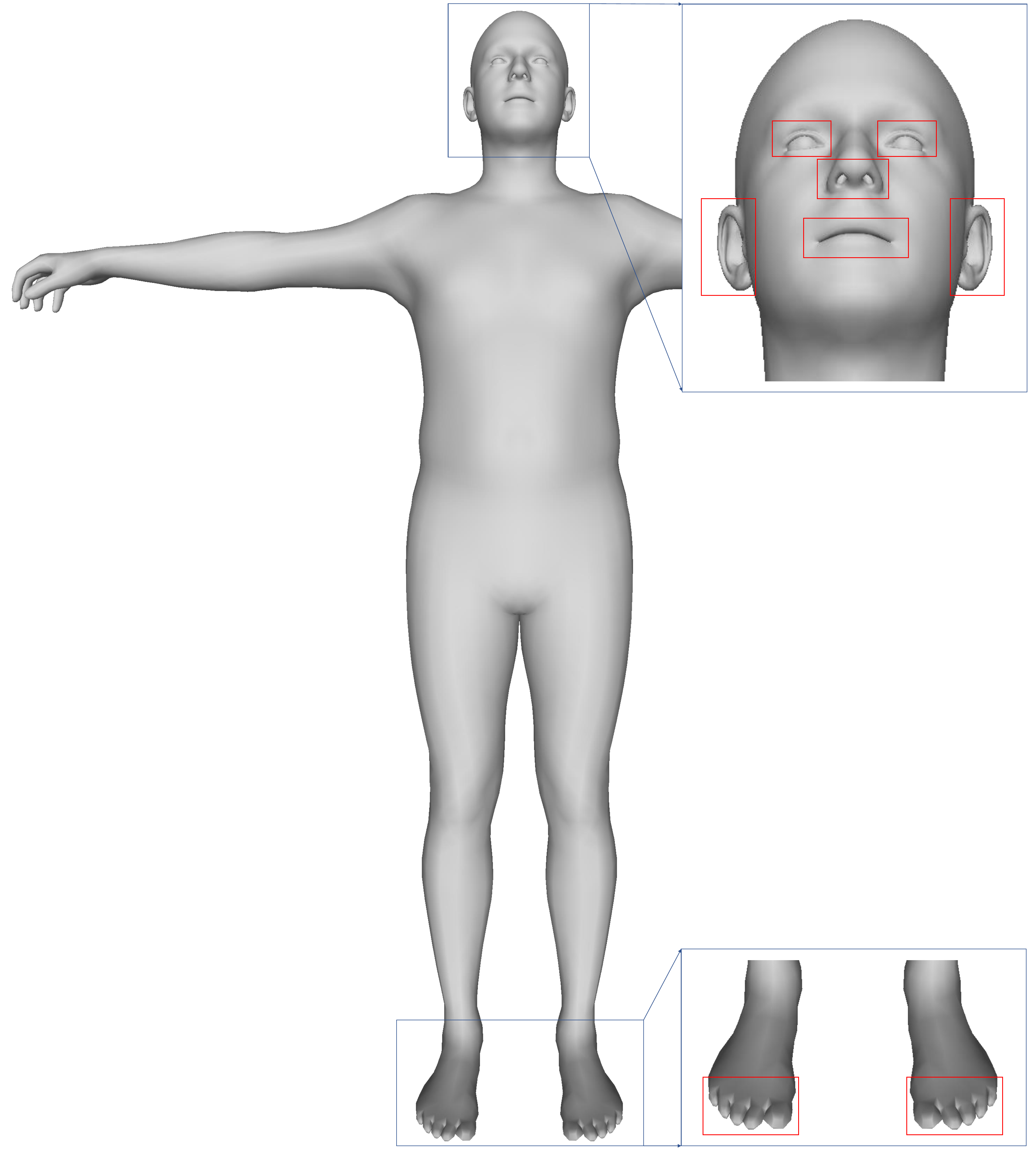}}
      \subfloat[Vertex reduction]{\label{fig:transfer2} \includegraphics[width=0.32\textwidth]{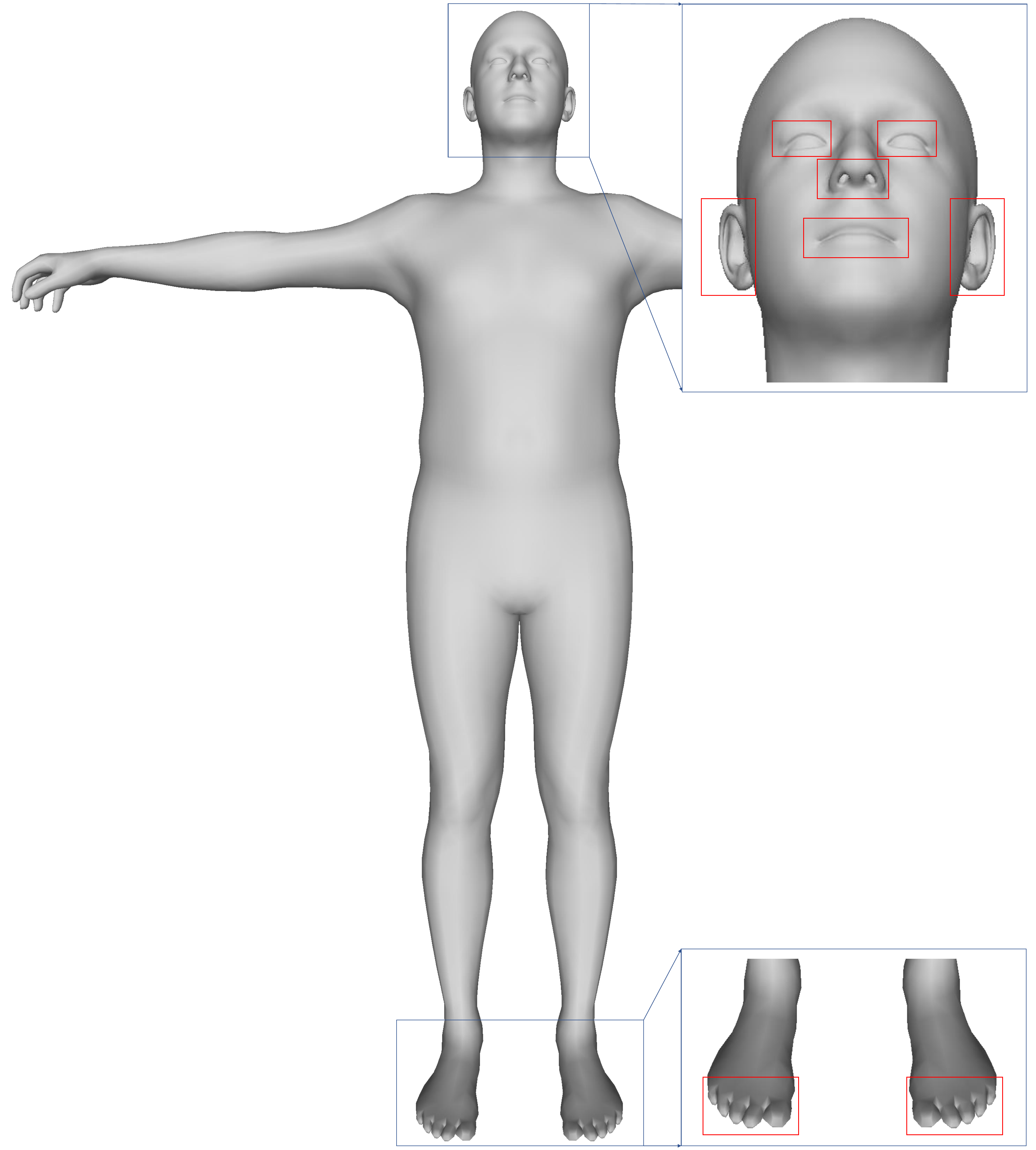}}
      \subfloat[Face flattening]{\label{fig:transfer3} \includegraphics[width=0.32\textwidth]{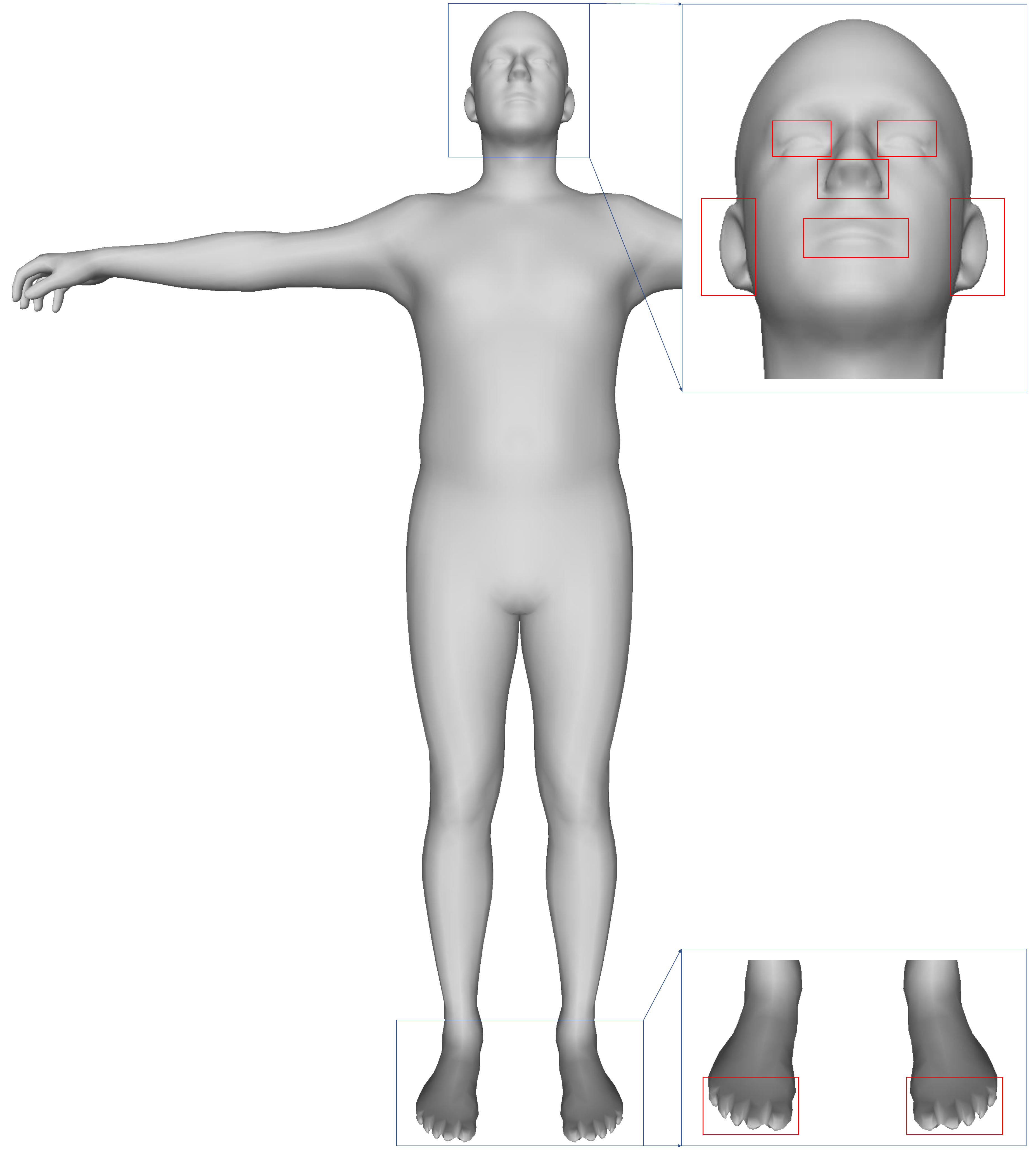}}
      \\
      \subfloat[Toe seam process]{\label{fig:transfer4} \includegraphics[width=0.32\textwidth]{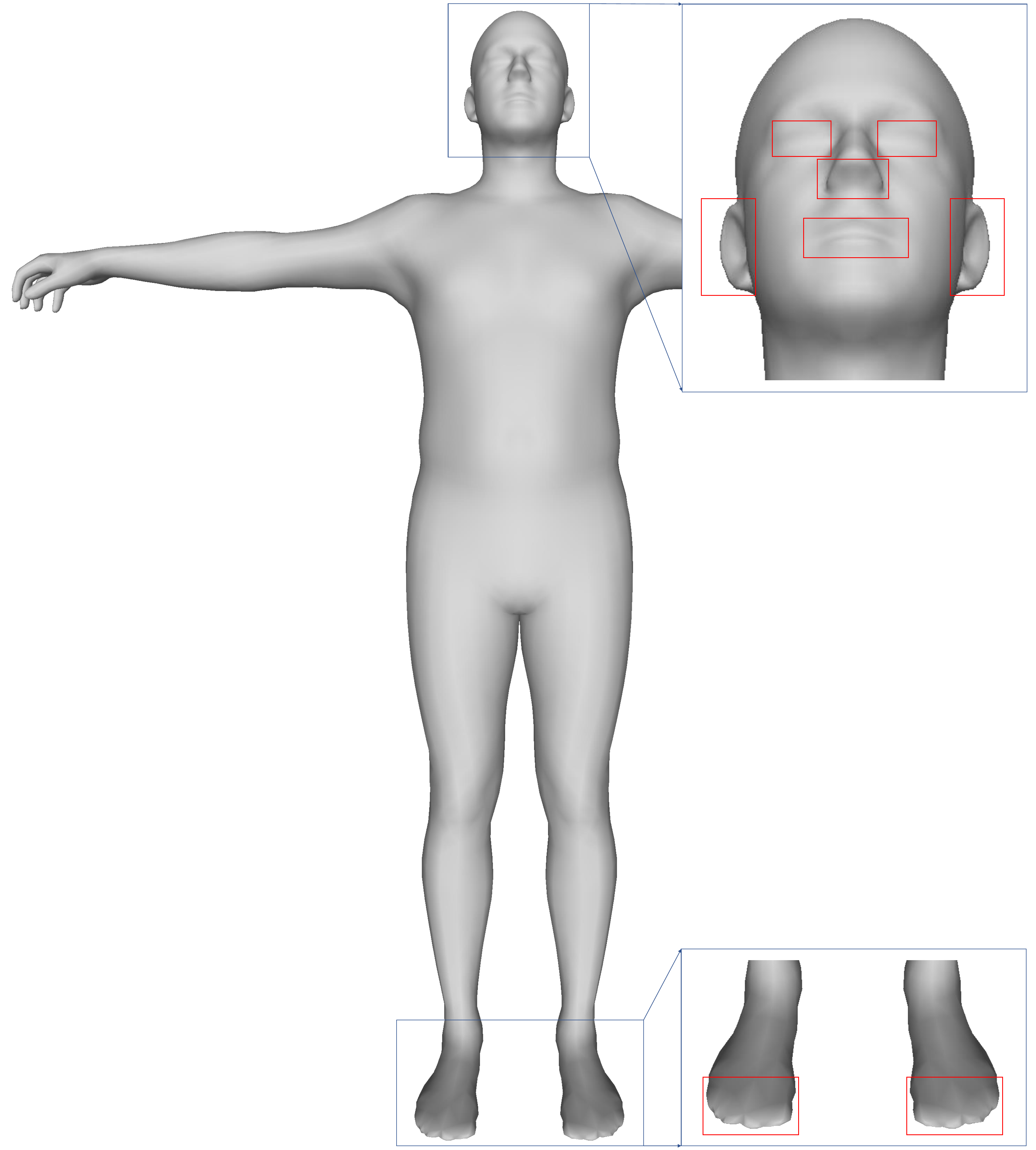}}
      \subfloat[SMPLX-Lite]{\label{fig:transfer5} \includegraphics[width=0.32\textwidth]{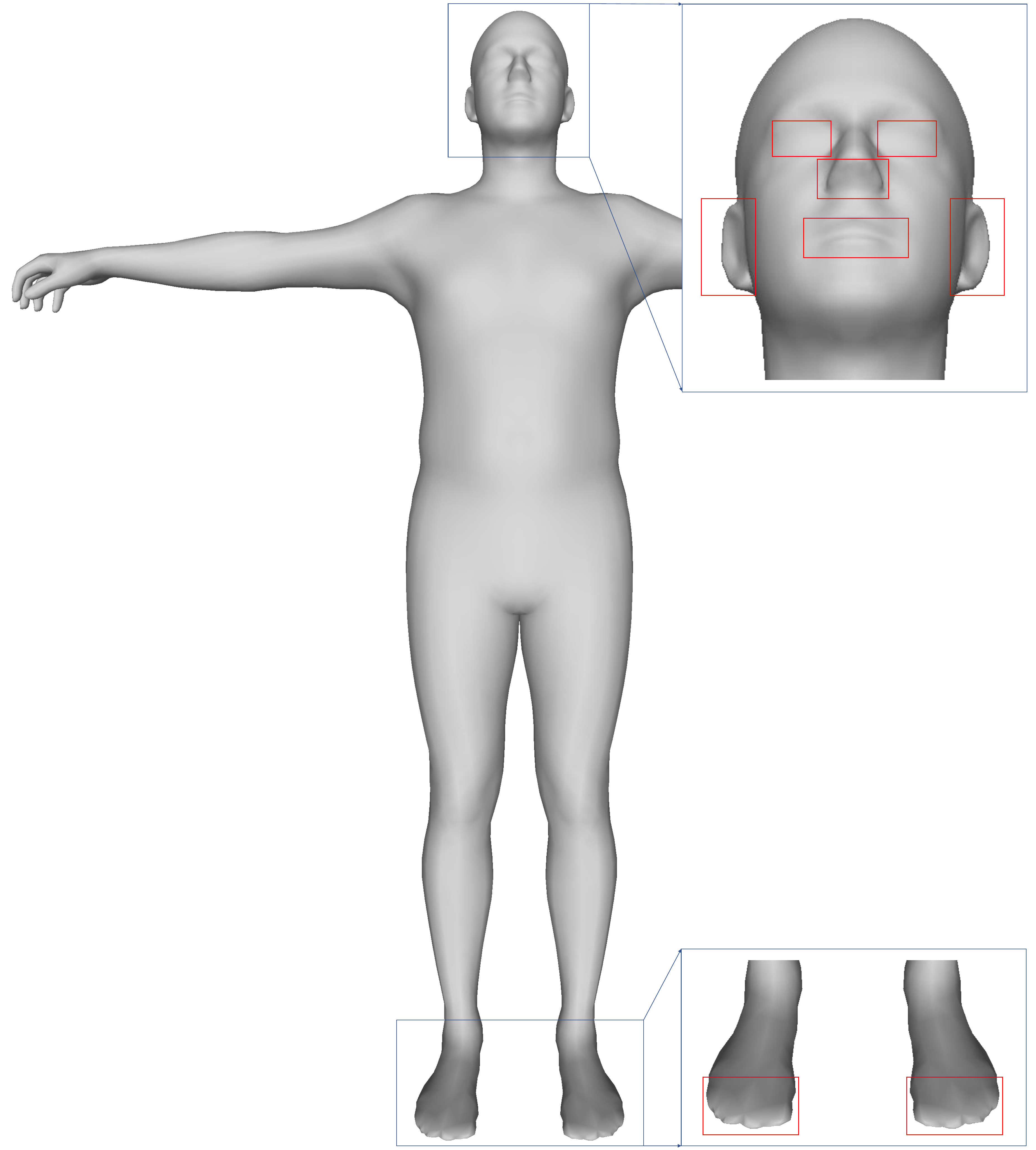}}
    \caption{\textbf{Model Transfer Process.} (a) SMPL-X model. (b) Intermediate model after vertex reduction. (c) Intermediate model after face flattening. (d) Intermediate model after toe seam process. (e) Final SMPLX-Lite model.}
    \label{fig:transfer}
\end{figure*}

We present the new SMPLX-Lite parametric model, which is derived from SMPL-X.
The model aims to capture the intricate geometry of the scanned mesh, while also ensuring stable geometry in critical areas such as the nose, mouth, and feet, as well as preserving the overall facial and finger shapes.
The entire process is depicted in Fig.\ref{fig:transfer}. 

First, we eliminate the vertices within the eyeballs, cochlea, lips, nostrils, and toe seam region from the SMPL-X (\ref{fig:transfer1}) model that are either hidden or folded.
Subsequently, we connect the edge vertices to create faces, while keeping the remaining vertices and topology unaltered. 
The resulting model (\ref{fig:transfer2}, \ref{fig:transfer3}) still has a large depression area, which could affect vertex fitting. 
Consequently, we flatten the faces in these particular regions to achieve a smoother surface, ensuring a uniform vertex distribution. 
Nonetheless, it is observed that the vertex and face distribution remains uneven during the fitting process, resulting in clustering of some vertices and severe distortion of corresponding faces (refer to Fig.\ref{fig:transfer4}).
To address this issue, we undertake multiple rounds of vertex deletion, face reconstruction, and face flattening to obtain a more suitable model for vertex fitting, which we designate as the SMPLX-Lite model (\ref{fig:transfer5}).

Subsequently, the reduction in the number of vertices necessitates adjustments to the matrices $\mathcal{S}$, $\mathcal{E}$, $\mathcal{P}$, $\mathcal{J}$, and $\mathcal{W}$, as described in Sec2.2 of the main paper to ensure that the transferred model inherits the control parameters of SMPL-X and the linear blend skinning function.
Initially, we resize these matrices to $\mathbb{R}^{N \times *}$, where $* \in \{3|\beta|, 3|\psi|, 3 \times 9K, K, K\}$, to ensure that the number of rows remains consistent across all matrices.
Then, for the $\mathcal{S}$, $\mathcal{E}$, $\mathcal{P}$, and $\mathcal{W}$ matrices, we determine the nearest neighbor on the SMPL-X model for each vertex of the SMPLX-Lite model, and uses the corresponding row in the original matrix to populate the new matrix. 
However, for the $\mathcal{J}$ matrix, using the nearest neighbor will result in a loss of regression coefficients for certain vertices to joints. 
To circumvent this, we identify the nearest neighbor on the SMPLX-Lite model for each vertex of the SMPL-X model, and subsequently aggregate the rows corresponding to the same point on the SMPL-X Lite model as a row of the new matrix.

\subsection{SMPLX-Lite-D fit}
\label{sec:fit}

We describe in detail the 2 stages of SMPLX-Lite-D fit process in Sec.2.2 of the main paper.

Stage 1: Embedded Nodes.  
The embedded nodes are initialized on the T-pose mesh without clustering vertices by radius as done in \cite{2009Robust}.
Instead, we cluster vertices by connectivity. 
The unbalanced distribution of embedded nodes is naturally adapted to the distortion ability of SMPLX-Lite mesh surface.
\begin{enumerate}[1)]
    \item We initialize a candidate set $S_{0}$ with all the vertices on the mesh. We randomly select $1$ vertex from the candidate set $S_{0}$ as a new embedded node and remove $k$ level of neighbor vertices from the candidate set, forming the remaining set $S_{1}$. By $k$ level of neighbors, we refer to at least $k$ jumps from the select vertex to the neighbor vertex. In practice, we use $k = 2$.
    \item Repeat step 1) until the candidate set is empty.
    \item For a embedded node $x_{i}$, we define a base radius $r_{i}$ as the average radius of its $k$ level of neighbors. We define the weight of a embedded node w.r.t. a vertex $v_{j}$ by their geodesic distance $d_{ij}$:
    $$
    w(v_{j}, r_{i}) = max(0, (1 - \frac{d_{ij}^2}{(\alpha  r_{i})^2})^3), 
    $$
    where $\alpha$ controls how far an embedded node can affect. 
    In practice, we have $\alpha = 1.5$.
    \item The smooth term between embedded nodes is defined in a similar way to step 3). 
    Two embedded nodes are considered neighbors if the geodesic distance between them is less than twice the largest base radius of them.
\end{enumerate}
Upon initializing the embedded nodes on T-pose mesh, we record the selected vertex indices and weights. 
When applied to a posed SMPLX-Lite model, the embedded nodes' positions are initialized using the corresponding vertex positions on the posed mesh. 
Please note that even the embedded nodes have the same initial positions as the chosen vertices, they are not bonded to be the same during the subsequent solving iterations.

To solve for the rotation and translation of the embedded nodes, our objective is to minimize the distance of warped vertices towards their nearest match on the scanned mesh. 
We refer the readers to \cite{2009Robust} for more details.

Stage 2: Vertex Shifts. After the fitting with embedded nodes, we only need to solve for tiny vertex shifts to ultimately capture the fine geometry details. 
With the final shifts regularized by a Laplacian matrix initialized from the resulting mesh of Stage 1, the fitted mesh is denoted by SMPLX-Lite-D.

\begin{figure}[ht]
    \centering
    \subfloat[Scanned\\ mesh]{\label{fig:scan} \includegraphics[width=0.23\linewidth]{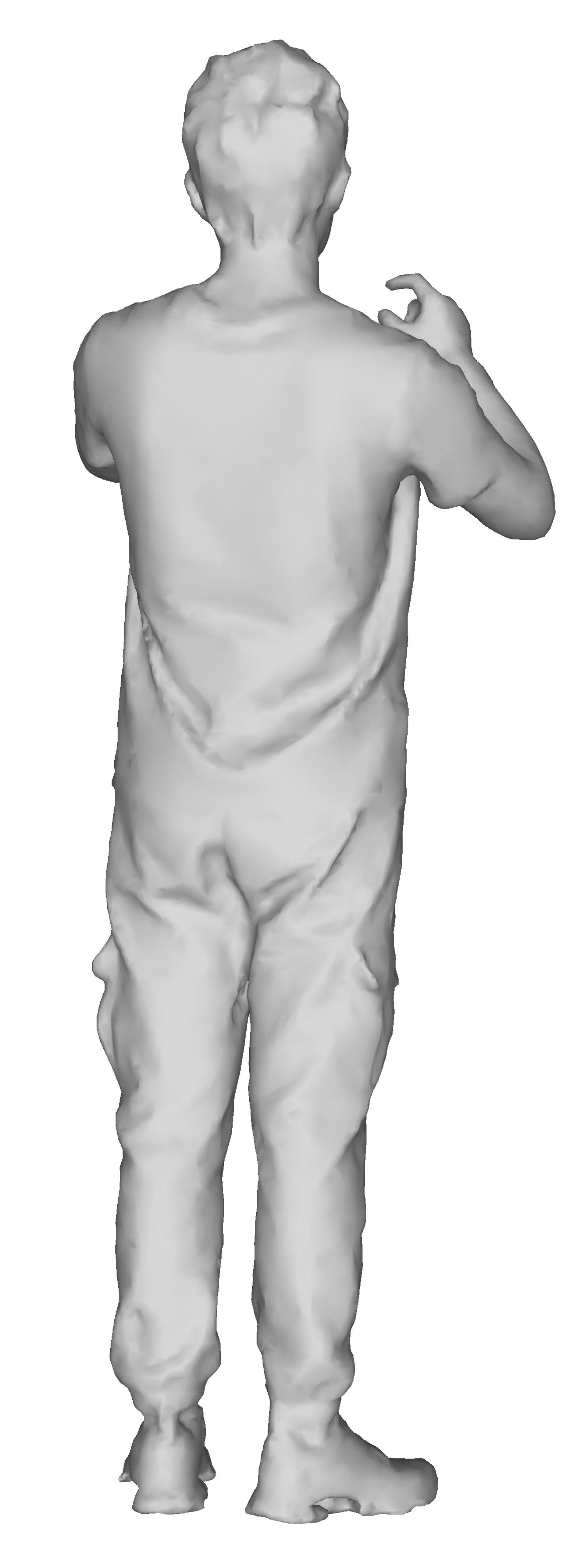}}
    \subfloat[Embedded\\ nodes]{\label{fig:Embedded nodes} \includegraphics[width=0.23\linewidth]{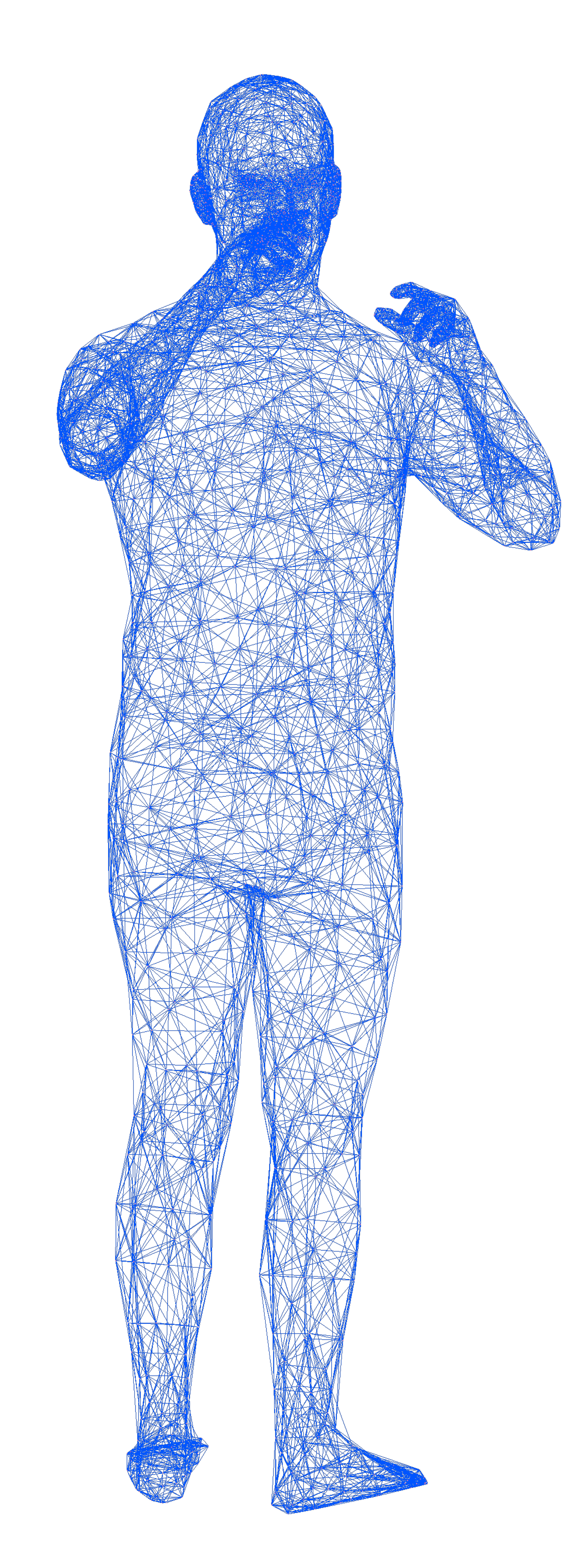}}
    \subfloat[Stage 1\\ results]{\label{fig:Stage 1} \includegraphics[width=0.23\linewidth]{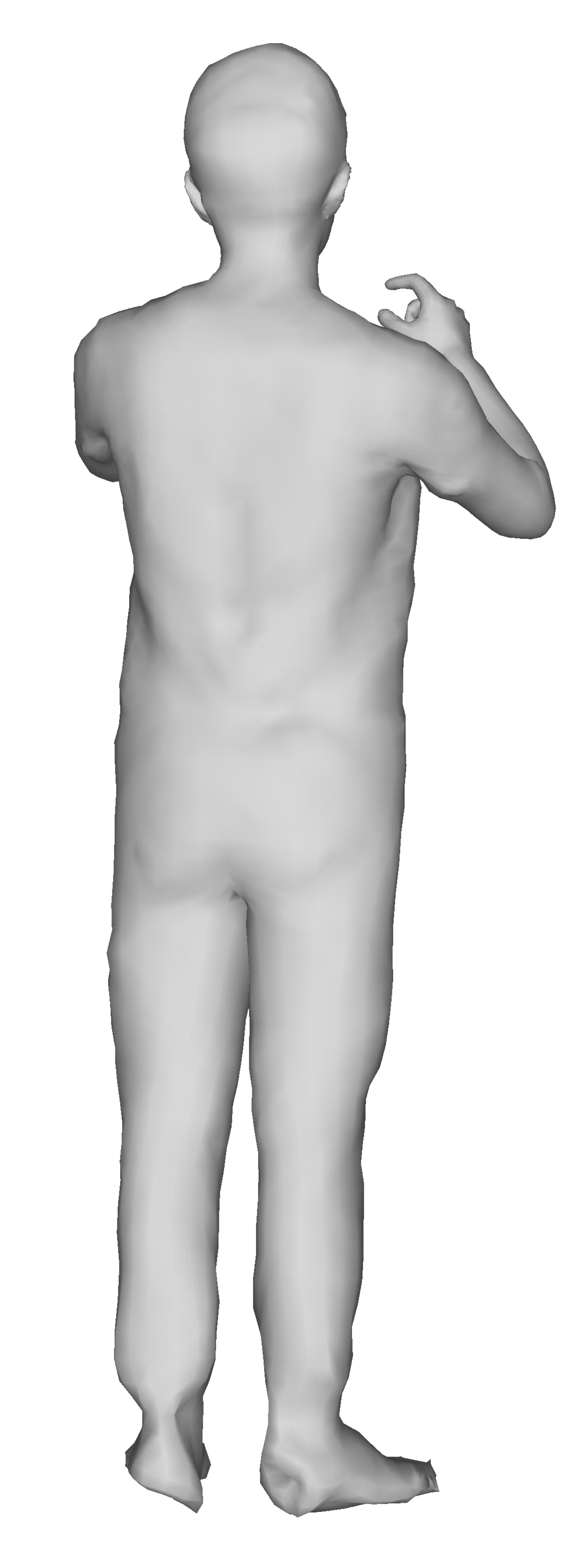}}
    \subfloat[Stage 2\\ results]{\label{fig:Stage 2} \includegraphics[width=0.23\linewidth]{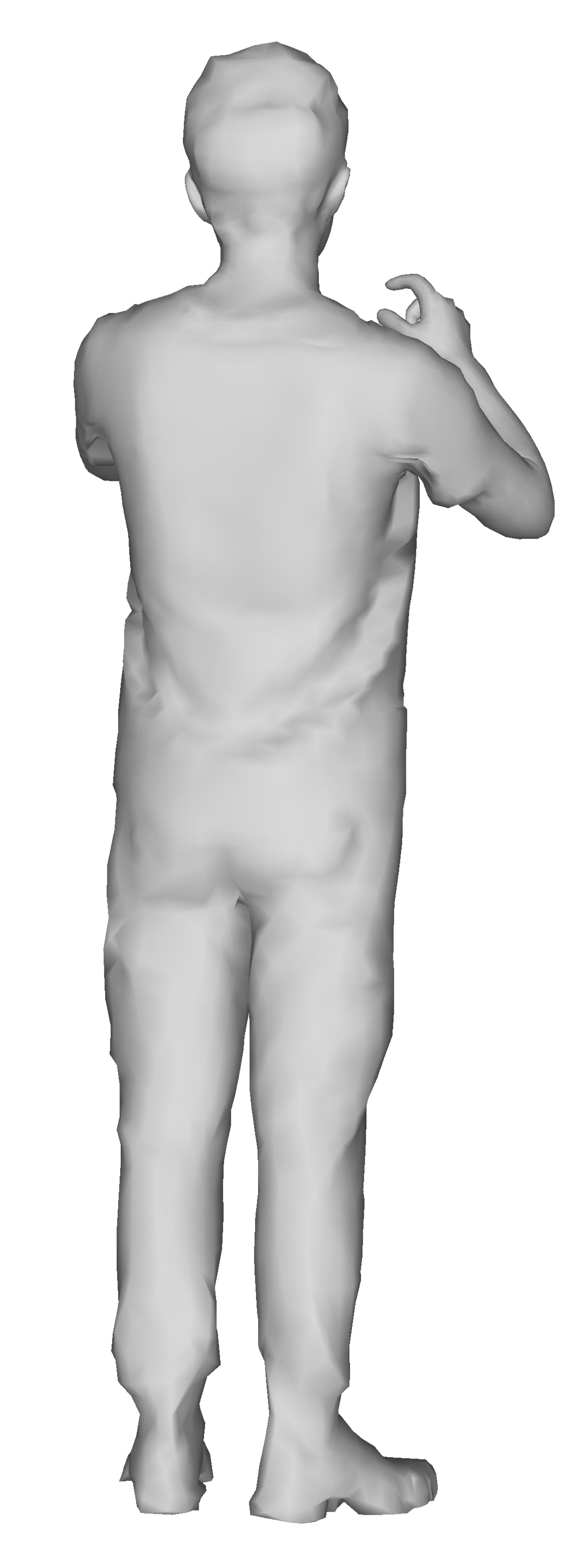}}
    \caption{\textbf{SMPLX-Lite-D Fit Pipeline.} (a) The scanned mesh. (b) The embedded nodes that control the warping in fitting stage 1. (c) The result of fitting stage 1, where joints are fitted and the surface presents a few details. (d) The final result of fitting the Stage 2. More geometry details can be seen in the hair and cloth wrinkles.}
    \label{fig:fit pipeline}
\end{figure}

We present our fit pipeline in Fig.\ref{fig:fit pipeline}. The stage 1 reasonably fits the scanned mesh, while the Stage 2 presents more geometry details.

\begin{figure}[ht]
    \centering

      
    \subfloat[Ear w/o Stage 1]{\label{fig:ear01} \includegraphics[width=0.48\linewidth]{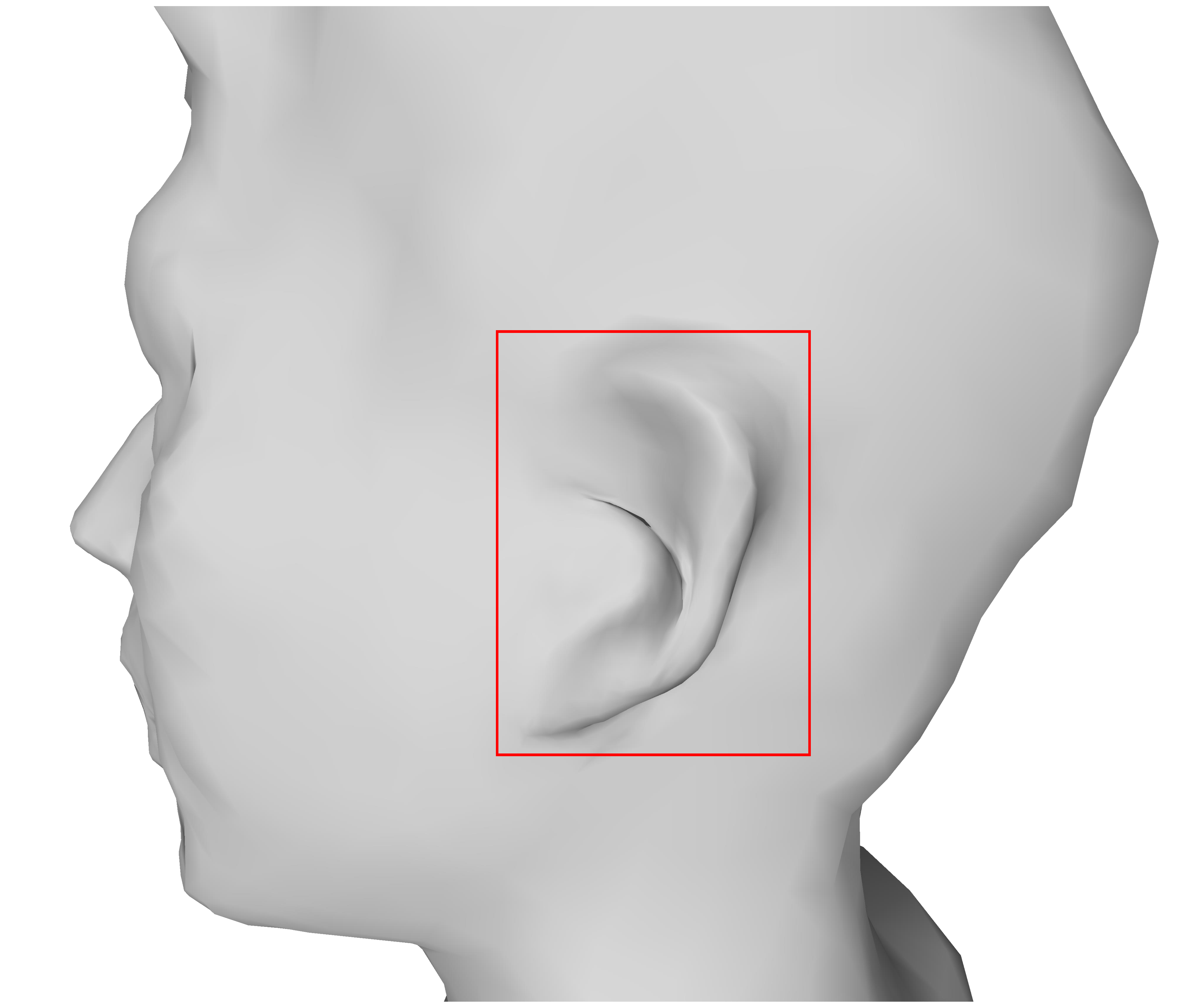}}
    \subfloat[Ear w/ Stage 1]{\label{fig:ear00} \includegraphics[width=0.48\linewidth]{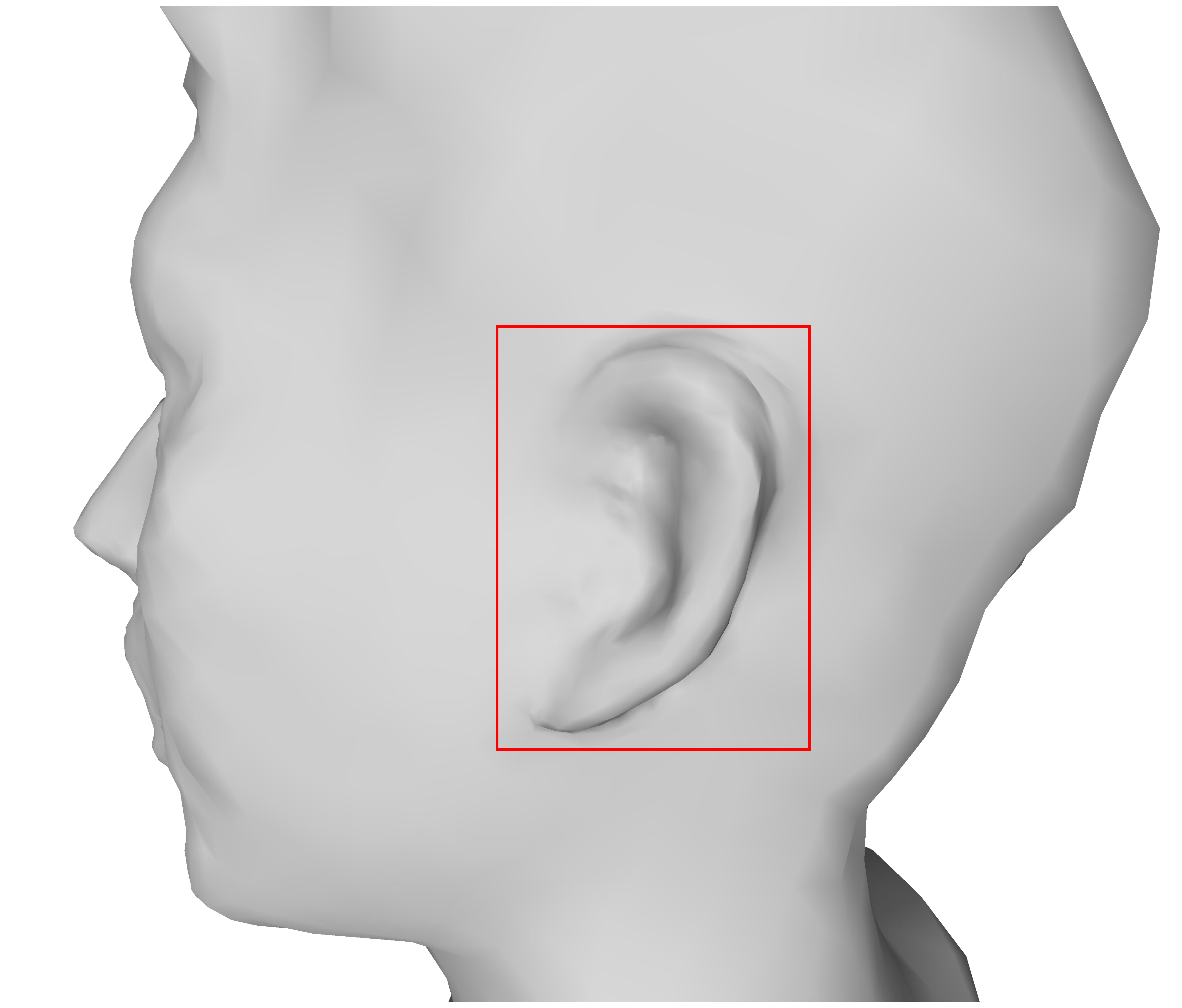}}
    \\
    \subfloat[Hand w/o Stage 1]{\label{fig:hand01} \includegraphics[width=0.48\linewidth]{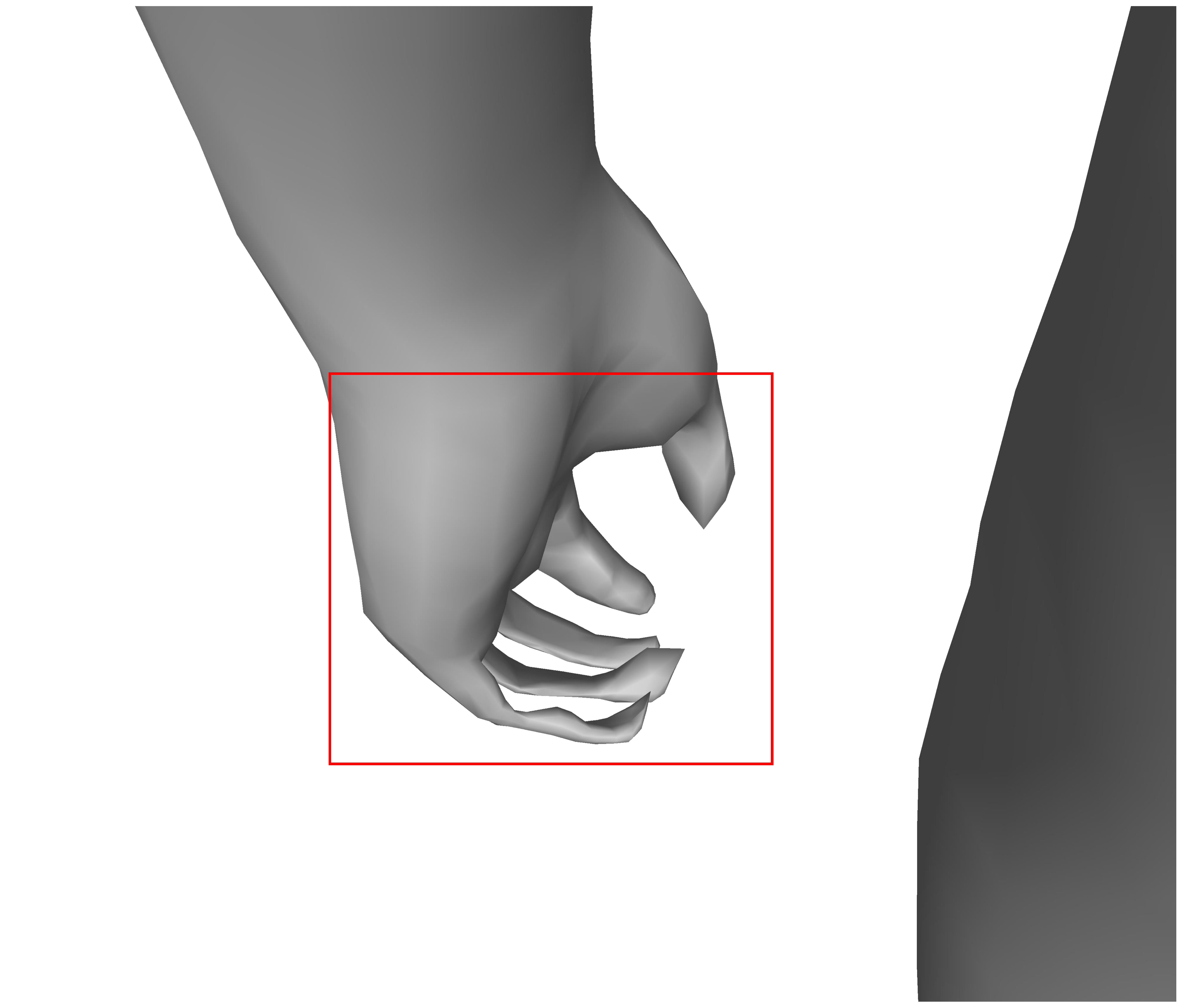}}
    \subfloat[Hand w/ Stage 1]{\label{fig:hand00} \includegraphics[width=0.48\linewidth]{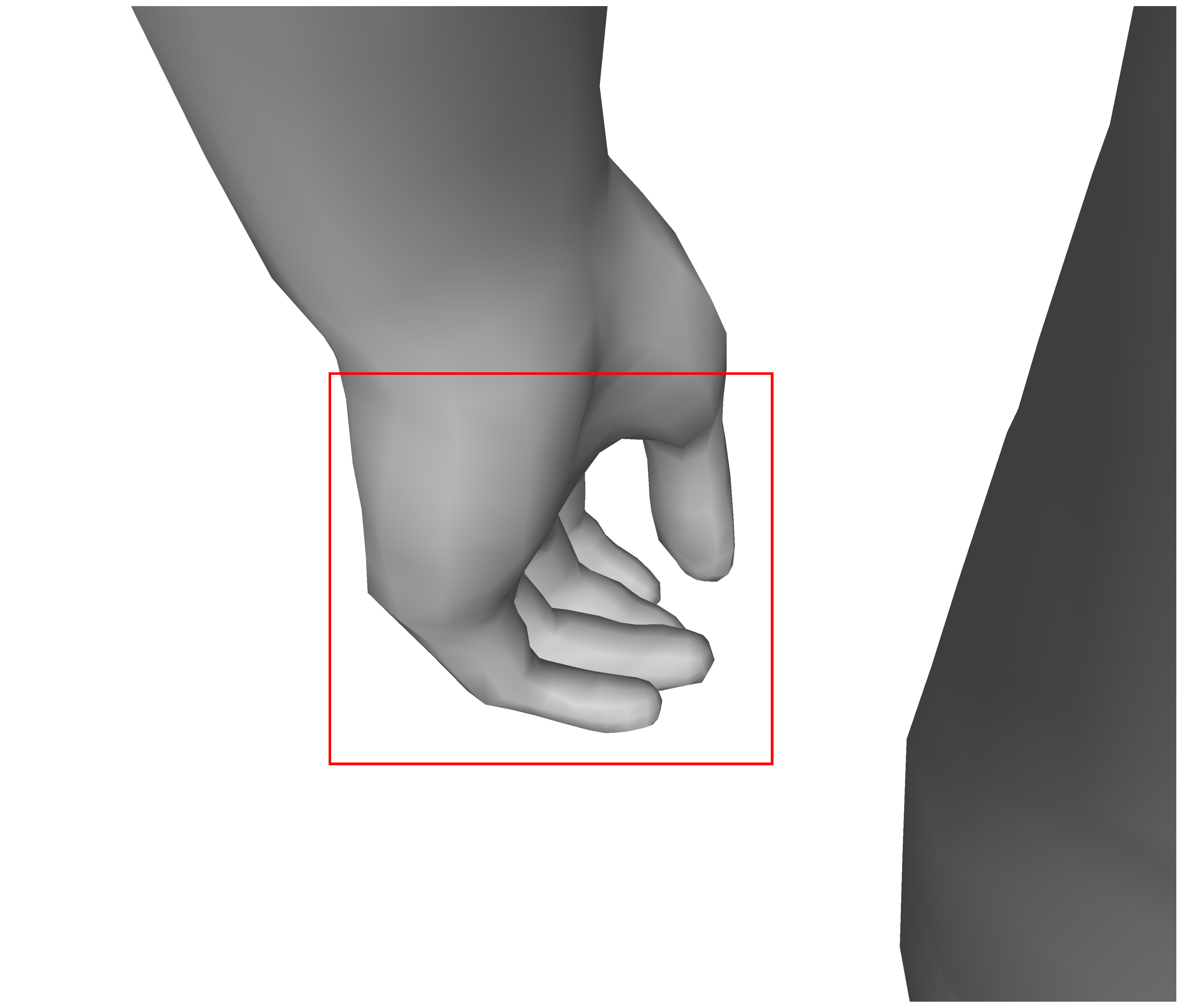}}
    
    \caption{\textbf{SMPLX-Lite-D Fit Comparison.} Large distortion is beyond the applicability of the laplacian matrix as a regularizer, leading to undesirable artifacts in the finger and ear areas, which are solved by Stage 1.}
    \label{fig:fit compare}
\end{figure}

We also compare the results of our two-stage fitting vs. direct vertex fitting (stage 2 only) in Fig.~\ref{fig:fit compare}. Direct vertex fitting may generate undesirable artifacts in many regions.

\section{Extended Dataset Evaluation Results}
\label{sec:eval}

\begin{figure*}[ht]
    \centering
      
    \subfloat[SMPL]{\includegraphics[width=0.18\linewidth]{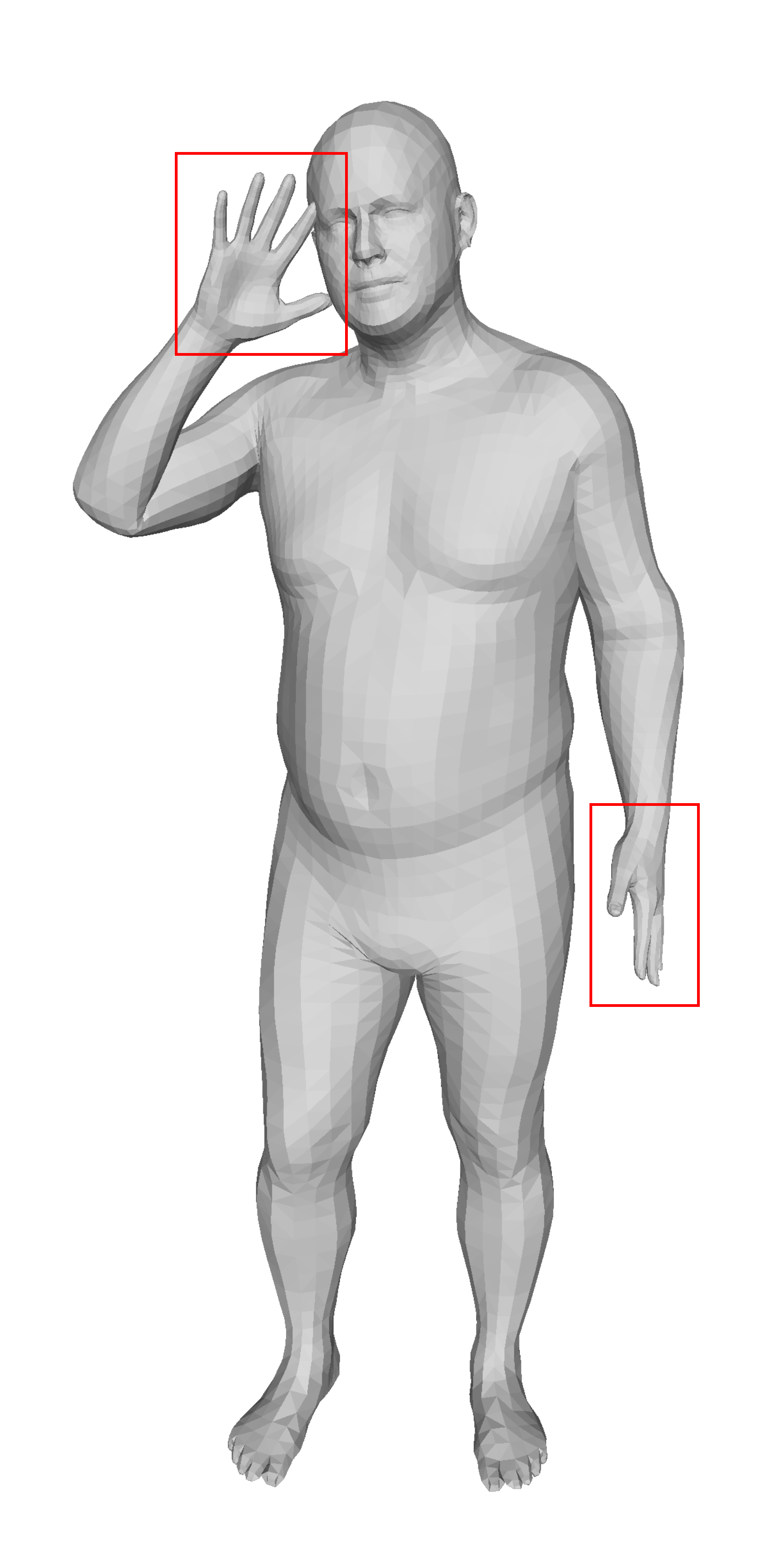}}
    \subfloat[SMPL-X]{\includegraphics[width=0.18\linewidth]{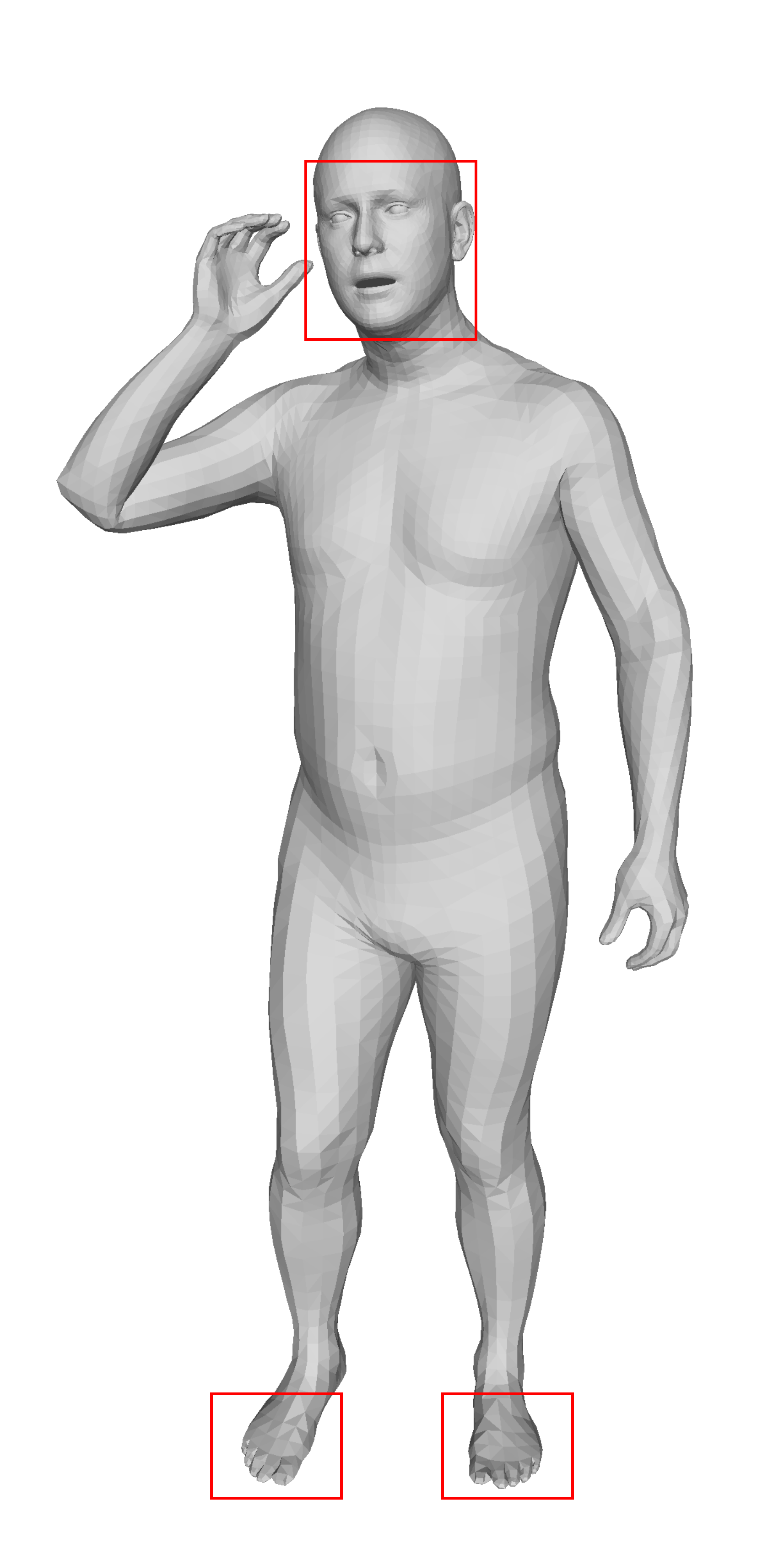}}
    \subfloat[SMPLX-Lite]{\includegraphics[width=0.18\linewidth]{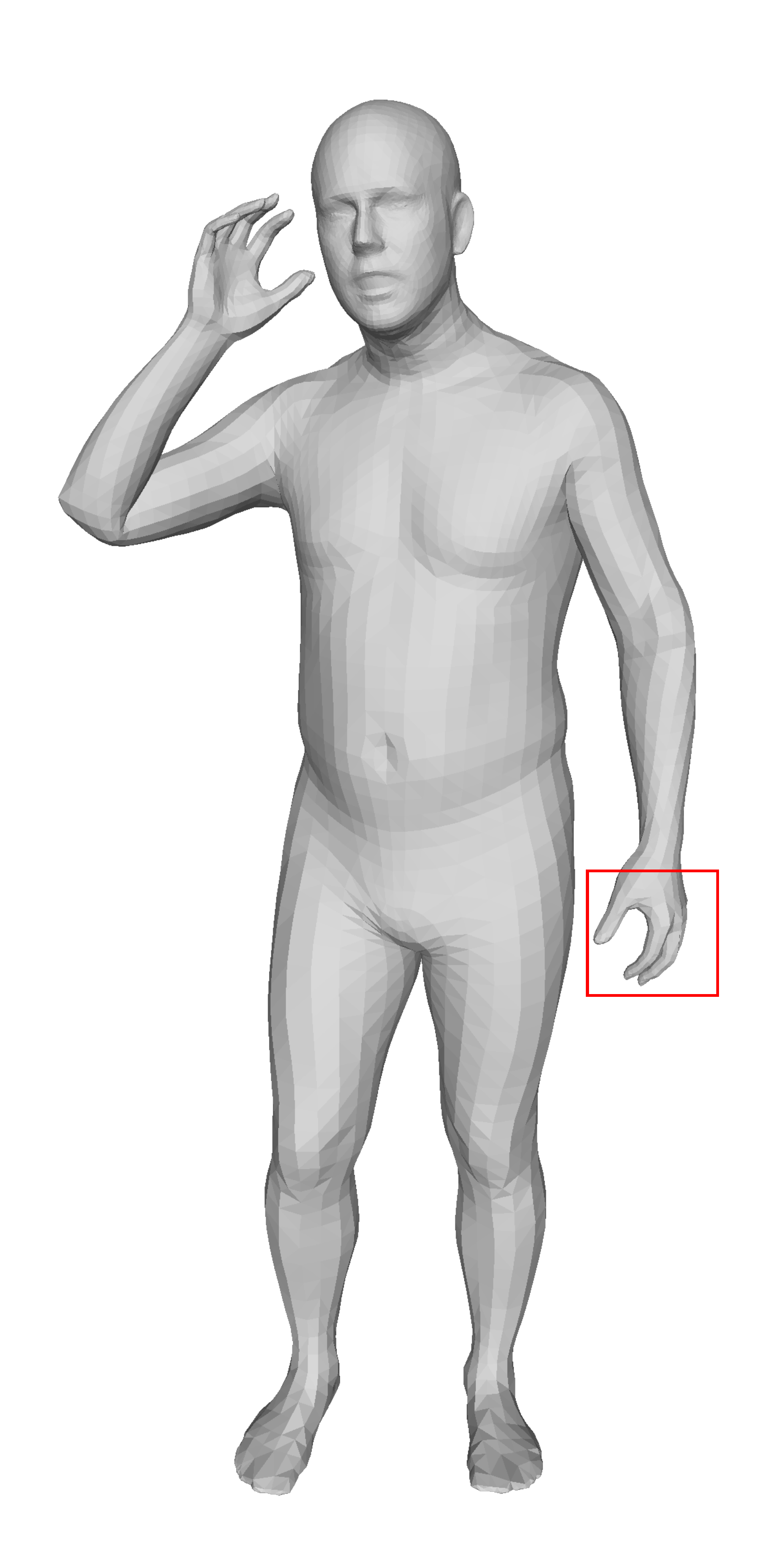}}
    \subfloat[SMPLX-Lite-D]{\includegraphics[width=0.18\linewidth]{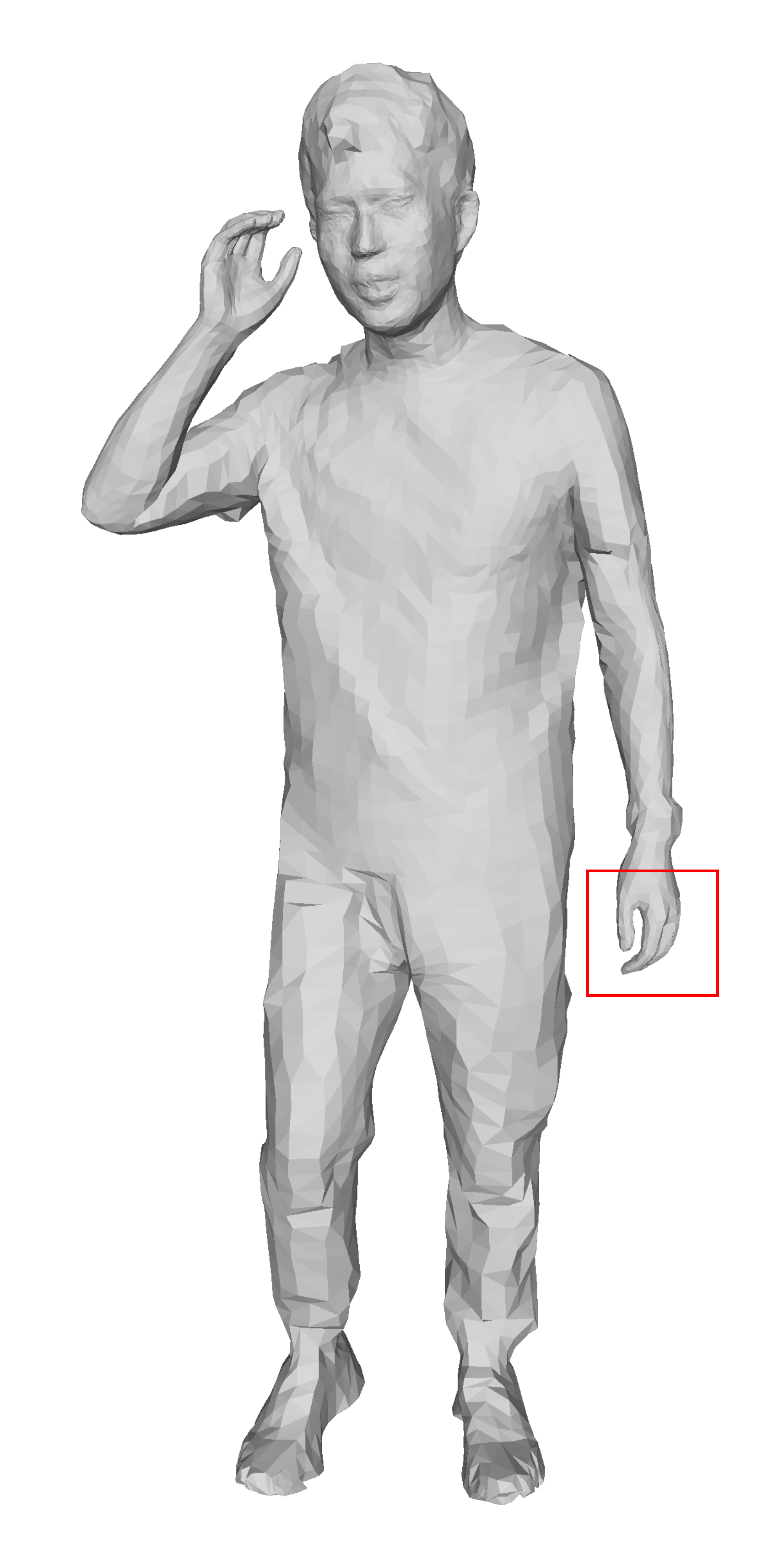}}
    \subfloat[Scanned Mesh]{\includegraphics[width=0.18\linewidth]{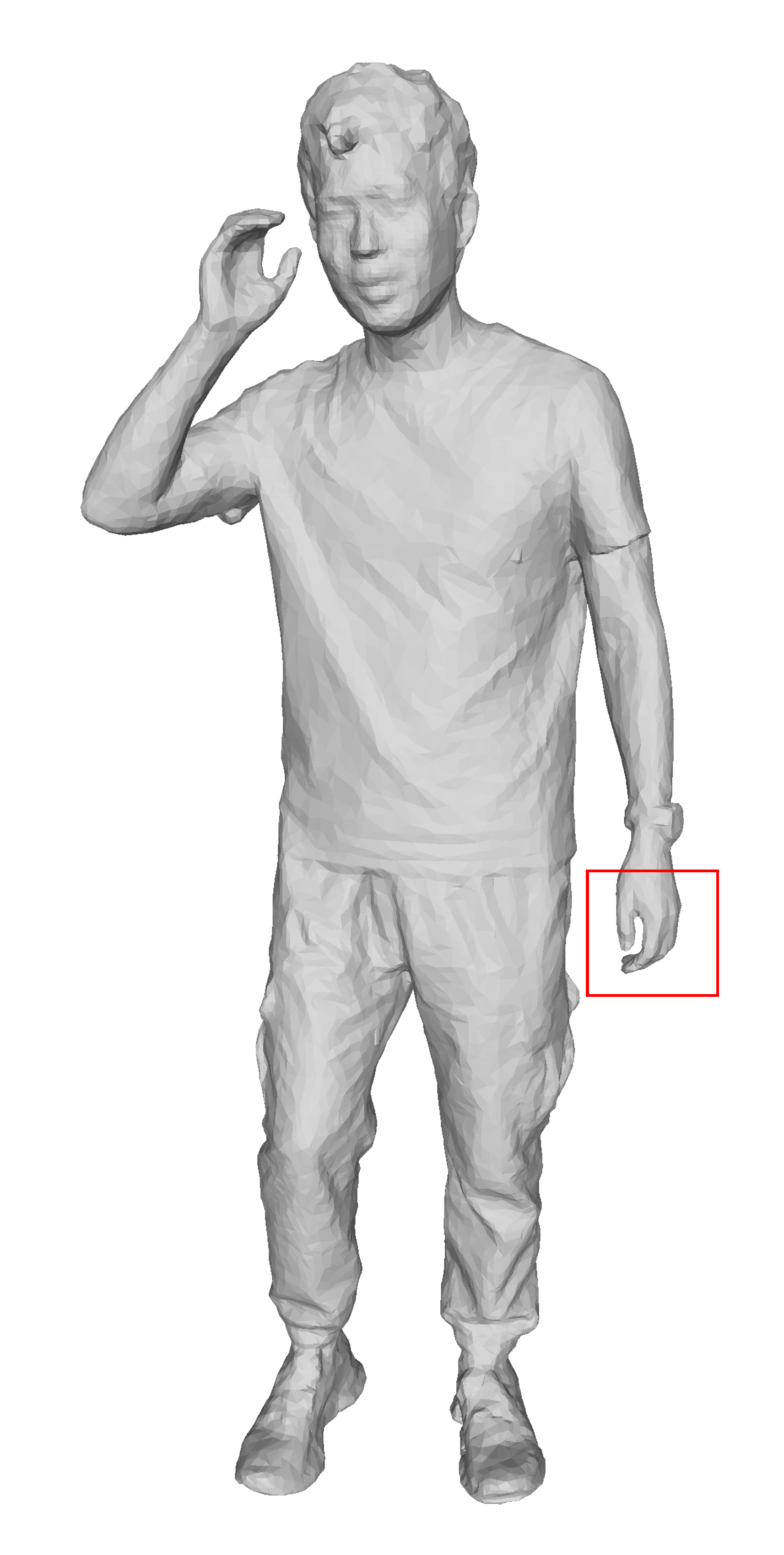}}
    
    \caption{\textbf{Fitting Results of Different Model} (a) SMPL model cannot control facial expressions and hand movements. (b) SMPL-X model has overly complex faces and toes, making it unsuitable for vertex fitting. (c) SMPLX-Lite model, plus vertex displacement (d) can fit scanned mesh(e) perfectly, especially in hand regions.}
    \label{fig:smpl compare}
\end{figure*}

\begin{figure*}[ht]
    \centering

    \newlength{\myheight}
    \settoheight{\myheight}{\subfloat[Recon]{\includegraphics[width=0.13\linewidth]{figures/reconstruction.pdf}}}
    \subfloat[Recon]{\includegraphics[height=\myheight]{figures/reconstruction.pdf}}
    \subfloat[Driving]{\includegraphics[height=\myheight]{figures/driving.pdf}}
    \subfloat[GT]{\includegraphics[height=\myheight]{figures/gt.pdf}}
    \subfloat[GT vs Recon]{\includegraphics[height=\myheight]{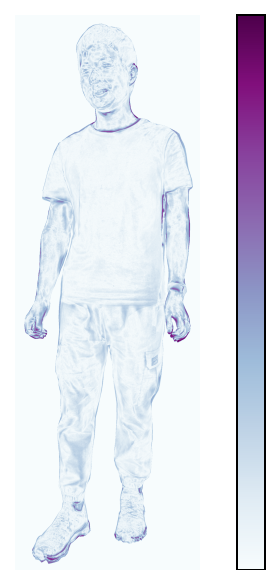}}
    \subfloat[GT vs Driving]{\includegraphics[height=\myheight]{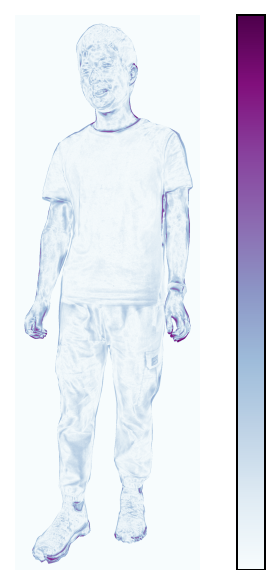}}
    \subfloat[Recon vs Driving]{\includegraphics[height=\myheight]{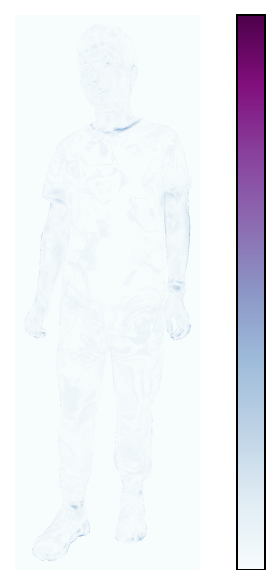}}
    
    \caption{\textbf{Qualitative Results and Difference Heatmap.} Both rendered images are really close to the captured image, perfectly recovering clothing details, finger movements, and facial expressions.}
    \label{fig:driving vs recon}
\end{figure*}

\begin{figure}[ht]
    \centering
    \includegraphics[width=0.4\textwidth]{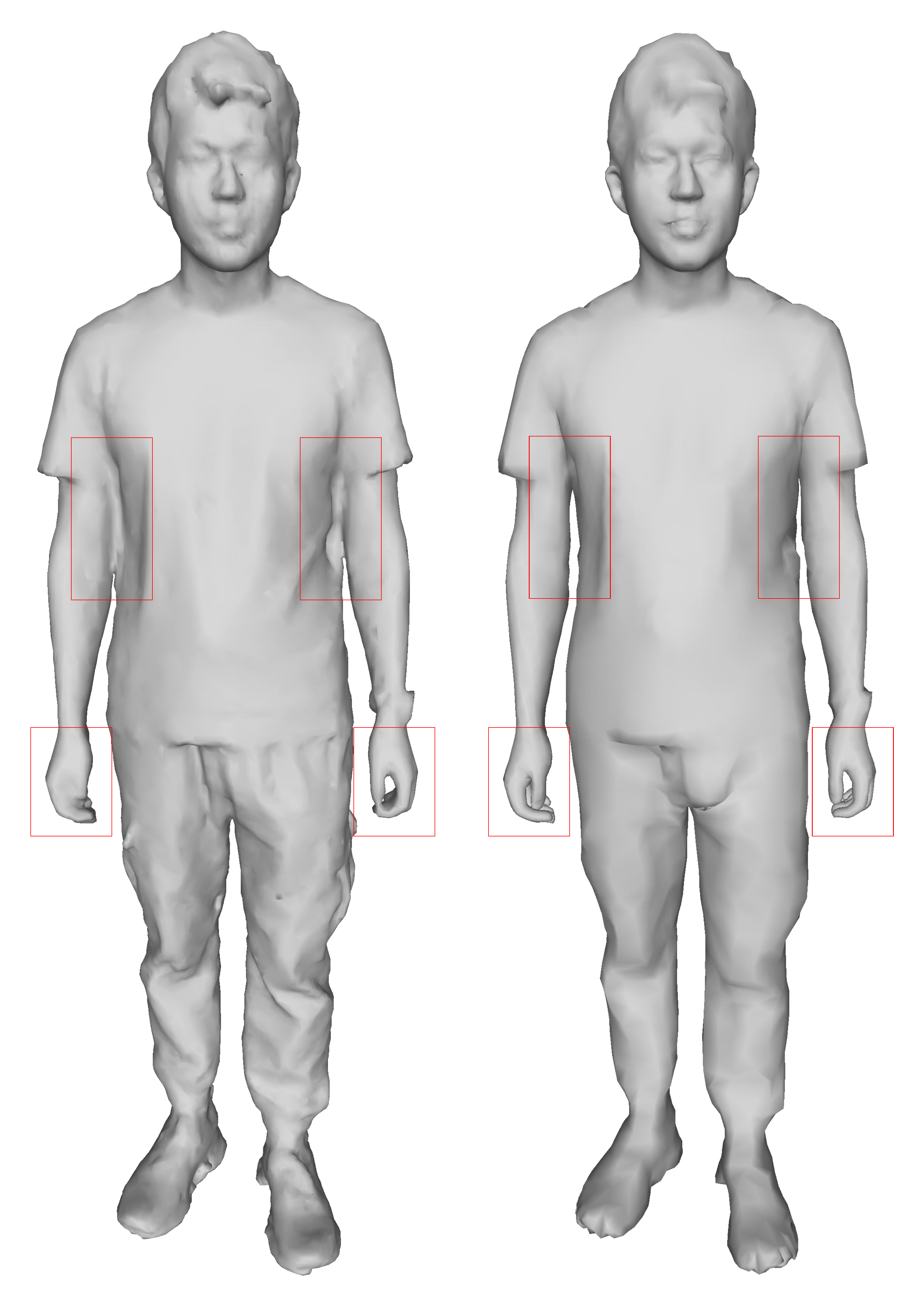}
    \caption{We show the adhesion in the hand and underarm areas of the scanned mesh (left one), and the SMPLX-Lite-D model (right one) does not have adhesion.}
    \label{fig:scan vs lite-d}
\end{figure}

The comparison of the SMPLX-Lite dataset with other datasets containing human model fits is presented in Tab.\ref{tab:dataset}.
As discussed in the main paper, SMPLX-Lite dataset offers a range of valuable components, including multi-view images, reconstructed texture models, and fitted clothed parametric models with texture maps. 
This variety of data types allows for the reconstruction of photorealistic drivable avatars, thereby providing researchers with a broader spectrum of supervising methods compared to datasets that only offer raw images \cite{cheng2022generalizable} or solely reconstructed textured models \cite{CAPE:CVPR:20}. 
In contrast, other datasets featuring both RGB images and scanned textured meshes are either synthetic or lack registered parametric models. Importantly, these datasets are unable to furnish a parametric model that facilitates control over facial expressions and hand movements and achieve vertex alignment. 
The fitting results of different parametric models are compared in Fig.\ref{fig:smpl compare}.
Notably, our registered SMPLX-Lite-D models enable multiple supervision methods, such as direct supervision of 3D mesh and texture, as well as supervision with 2D images.

\newcommand{\tabincell}[2]{\begin{tabular}{@{}#1@{}}#2\end{tabular}}

\begin{table*}
  \centering
  \caption{Comparison with existing datasets containing human model fits. SMPLX-Lite has multiple data types and annotations, and supports multiple tasks. Registered: parametric model fit; Vertex fit: parametric model can fit clothes or not; K3D: 3D keypoints; Act: action label; Sequence: sequential data. "Facebook" means the data used in \cite{2021Driving, 2021Modeling}, which is not public available.}
  \begin{tabular}{lccccccccc}
    \toprule
    Dataset      &     RGB      &     Mesh     &   Texture    &  Registered  &  \tabincell{c}{Vertex\\Fit}  & \tabincell{c}{Large Pose\\Variation} &     K3D      &     Act      &   Sequence   \\ \midrule
    RenderPeople\cite{RenderPeople} & $\checkmark$ & $\checkmark$ & $\checkmark$ &   $\times$   &   $\times$   &     $\checkmark$     & $\checkmark$ &   $\times$   & $\checkmark$ \\
  	DFAUST\cite{dfaust:CVPR:2017}       & $\checkmark$ & $\checkmark$ & $\checkmark$ &     Dyna     &   $\times$   &     $\checkmark$     & $\checkmark$ & $\checkmark$ & $\checkmark$ \\
  	BUFF\cite{Zhang_2017_BUFF}        & $\checkmark$ & $\checkmark$ & $\checkmark$ &     SMPL     &   $\times$   &     $\checkmark$     & $\checkmark$ & $\checkmark$ & $\checkmark$ \\
  	AGORA\cite{Patel:AGORA:2021}        & $\checkmark$ & $\checkmark$ & $\checkmark$ & SMPL-X\&SMPL &   $\times$   &       $\times$       & $\checkmark$ &   $\times$   &   $\times$   \\
  	HUMBI\cite{Yu_2020_HUMBI}       & $\checkmark$ & $\checkmark$ & $\checkmark$ &     SMPL     &   $\times$   &     $\checkmark$     & $\checkmark$ &   $\times$   & $\checkmark$ \\
  	THuman2.0\cite{tao2021function4d}    &   $\times$   & $\checkmark$ & $\checkmark$ &    SMPL-X    &   $\times$   &     $\checkmark$     &   $\times$   &   $\times$   &   $\times$   \\
  	ZJU LightStage\cite{peng2021neural}& $\checkmark$ & $\checkmark$ & $\checkmark$ &    SMPL-X    &   $\times$   &     $\checkmark$     & $\checkmark$ & $\checkmark$ & $\checkmark$ \\
  	GeneBody\cite{cheng2022generalizable}     & $\checkmark$ &   $\times$   &   $\times$   &    SMPL-X    &   $\times$   &     $\checkmark$     &   $\times$   &   $\times$   & $\checkmark$ \\
  	HuMMan\cite{cai2022humman}     & $\checkmark$ &   $\checkmark$   &   $\checkmark$   &    SMPL    &   $\times$   &     $\checkmark$     &   $\checkmark$  &   $\checkmark$  & $\checkmark$ \\
  	Sizer\cite{tiwari20sizer}        & $\checkmark$ & $\checkmark$ & $\checkmark$ &    SMPL-G    & $\checkmark$ &       $\times$       &   $\times$   &   $\times$   &   $\times$   \\
  	CAPE\cite{CAPE:CVPR:20}         &   $\times$   & $\checkmark$ &   $\times$   &    SMPL-D    & $\checkmark$ &     $\checkmark$     & $\checkmark$ & $\checkmark$ & $\checkmark$ \\
  	Facebook\dag\cite{2021Driving, 2021Modeling}     & $\checkmark$ & $\checkmark$ & $\checkmark$ & $\checkmark$ & $\checkmark$ &       $\times$       & $\checkmark$ &   $\times$   & $\checkmark$ \\
    \cmidrule(){1-10}
    Ours         & $\checkmark$ & $\checkmark$ & $\checkmark$ & SMPLX-Lite-D & $\checkmark$ &     $\checkmark$     & $\checkmark$ & $\checkmark$ & $\checkmark$ \\
    \bottomrule
  \end{tabular}
  
  \label{tab:dataset}
\end{table*}

Then, we provide more detailed dataset evaluation results.
We utilize 8 telephoto cameras and 24 standard cameras to capture images  with full body and local details simultaneously. 
PSNR and SSIM results of telephoto cameras are lower than standard cameras because they capture finer images, as shown in Fig.\ref{fig:dataset vis} and Tab.\ref{tab:eval}.

In Tab.\ref{tab:eval}, we have a complete list of the average results per act for each subject.
The names of the 15 actions are ``01 discussion", ``02 debating", ``03 presentation", ``04 eating", ``05 directions", ``06 greeting", ``07 purchasing", ``08 posing", ``09 waiting", ``10 walking", ``11 walking dog", ``12 phoning", ``13 taking photo", ``14 turning around", ``15 stretching". Some of the actions refer to the paper\cite{h36m_pami}.

\section{Extended Diverse Dataset Visualization}
\label{sec:vis}
\begin{figure*}[ht]
    \centering
    \includegraphics[width=0.9\textwidth]{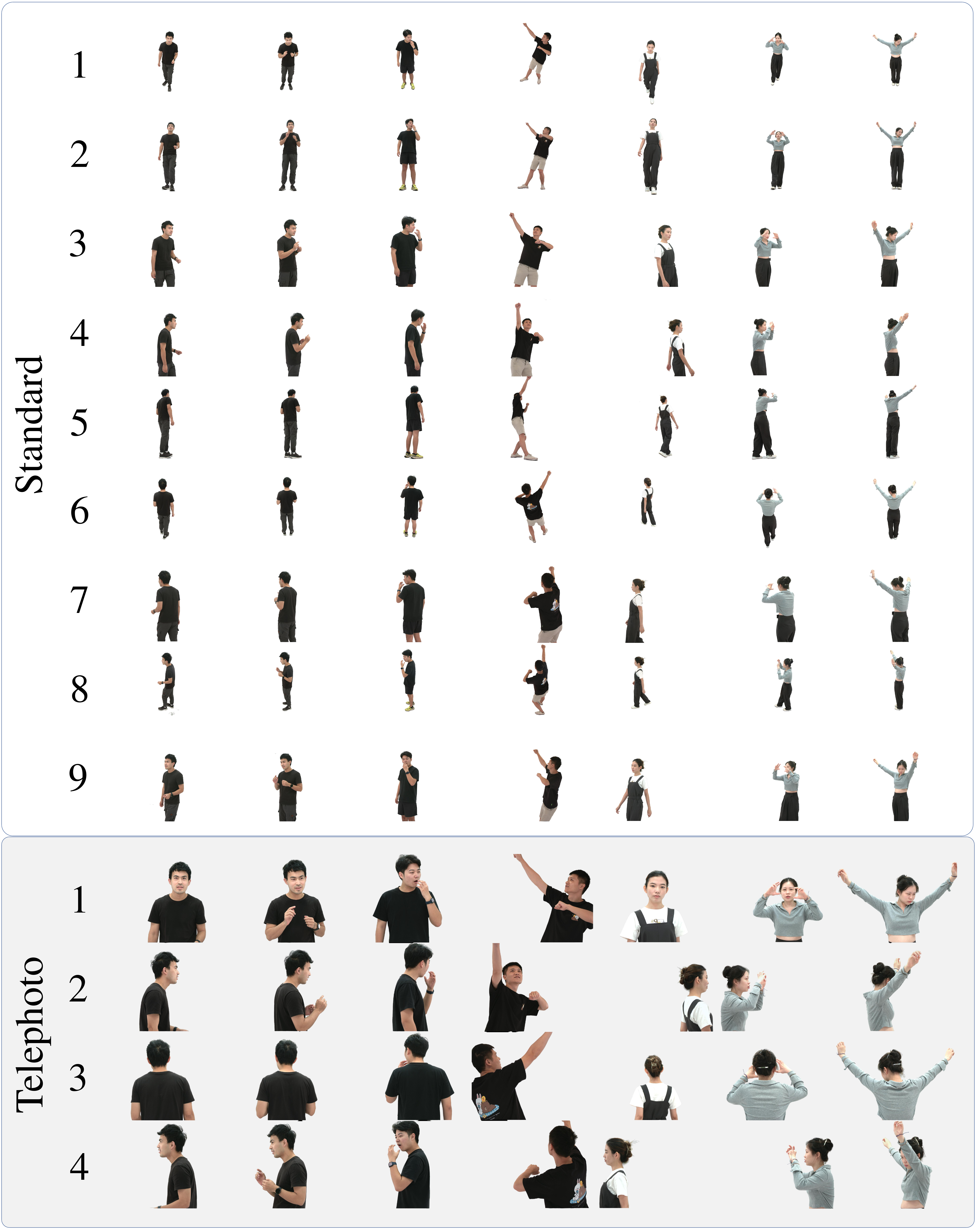}
    \caption{\textbf{Multi-View Capture.} SMPLX-Lite deploys 24 standard cameras and 8 telephoto cameras to capture multi-view synchronized RGB sequences. We show several frames of images from a part of these cameras.}
    \label{fig:dataset vis}
\end{figure*}

We present multi-view visualization in Fig.\ref{fig:dataset vis} and reconstructed high-resolution scan models of 5 subjects in Fig.\ref{fig:all}.

\section{Extended Experiments Results}
\label{sec:experiments_sup}

Our experiment settings are as follows: $epoch = 5$, $batch\_size = 1, lr = 5e-4, \lambda_{G} = 0.5, \lambda_{T} = 5, \lambda_{lap} = 50, \lambda_{KL} = 1$.
We utilize AdamW as optimizer and ExponentialLR as a scheduler with $\gamma = 0.9$

\subsection{Driving Results}
\label{sec:driving}
\begin{figure*}[ht]
    \centering
    \includegraphics[width=0.75\textwidth]{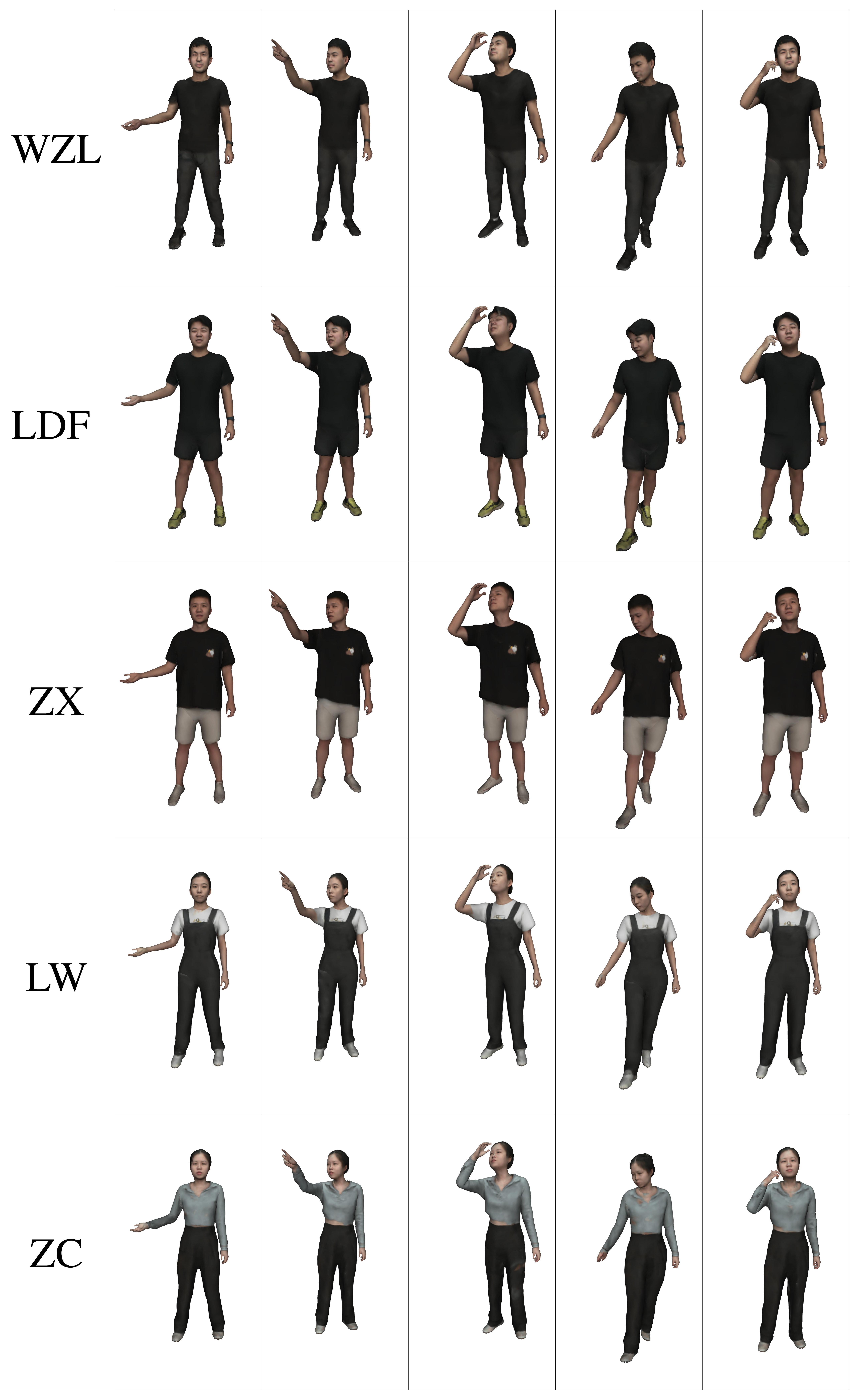}
    \caption{\textbf{Driving results of 5 models by the same driving signal}. Each column represents a different driving signal.}
    \label{fig:driving results_sup}
\end{figure*}

We train a drivable model for each subject and use the same driving signal to drive all the models, as shown in Fig.\ref{fig:driving results_sup}. 
Driven by the same signal, all reconstructed human models can present corresponding actions and facial expressions, and the geometry and texture of clothes change reasonably with the change of pose.

\subsection{Driving vs. Reconstruction}
\label{sec:compare}
We further visualize the qualitative results in Fig.5 of the main paper in Fig.\ref{fig:driving vs recon}. 
We nonlinearly transform the difference between every two images and get heat maps.
From the heatmap, we can see that the driving results are very close to the reconstruction results and both restore the captured image, perfectly recovering clothing details, finger movements, and facial expressions.

\subsection{Ablation Study}
\label{sec:ablation}

We perform ablation experiments to compare the effects of texture and image supervising.
The experiment settings are as follows: Tex: $\lambda_{G} = 0.5, \lambda_{T} = 5$; Img: $\lambda_{G} = 0.5, \lambda_{I} = 5, \lambda_{M} = 5$; Both: $\lambda_{G} = 0.5, \lambda_{T} = 2.5, \lambda_{I} = 2.5, \lambda_{M} = 2.5$.
As results in Tab.\ref{tab:ablation} demonstrate, texture map supervising works better than image supervising. The result of using both to supervise and simply averaging loss weights is the worst.

\begin{table}
  \caption{\textbf{Ablation Study Results.} We use either texture loss or image loss to supervise the generated texture map, and we also test the results of using both. The best results can be attained by using texture map supervision, which is only possible with our dataset.}
  \label{tab:ablation}
  \centering
  \begin{tabular}{lccc}
    \toprule
    Supervise & PSNR$\uparrow$ & SSIM$\uparrow$ & CD$\downarrow$($\times 10^{-3}$)\\
    \midrule
    Texture & \textbf{26.17} &\textbf{0.9396} &\textbf{4.5589}  \\
    Image & 26.01&0.9335 &9.0415 \\
    Both & 19.32 &0.5925  &40.000 \\
    \bottomrule
  \end{tabular}
  
\end{table}


\section{Discussion of Limitations}
\label{sec:future work}
In this section, we discuss several limitations of the SMPLX-Lite dataset and driving method. 

As shown in Fig.~\ref{fig:scan vs lite-d}, the scanned mesh reconstructed from the depth map and the point cloud has adhesion in very close areas, such as hands and underarms, while the fitted parametric model SMPLX-Lite-D has not. 
Therefore, chamfer distance (CD) may not be the most appropriate evaluation metric and does not reflect the advantages of our fitted model.
A more reasonable evaluation metric is needed to evaluate the quality of the fitted mesh.

As for the driving method, our proposed one is only a preliminary baseline, which works well overall, but artifacts can occur when driving out-of-distribution actions. 
Besides, the current algorithm is still elementary for facial expression control. 
To get a drivable model with good generalization capabilities, a large amount of data is needed to train the neural network, which our dataset now provides. 

In future studies, we will further promote the diversity and number of action sequences and optimize the SMPLX-Lite-D fit results. 
We will improve the baseline driving algorithm to take full advantage of the diverse data in the SMPLX-Lite dataset to achieve a better driving effect and consider decoupling of expressions and whole-body poses to produce more vivid facial expressions.
Also, we consider using fewer data to train available models and achieve training time reduction.


\clearpage
\onecolumn
\begin{longtable}{l|c|ccccccccc}
	\caption{\textbf{Complete Dataset Evaluation Results.}We render textured models of 5 subjects to 32 views (8 telephoto cameras and 24 standard cameras), compare them with captured images to get PSNR and SSIM and compare the geometry of the fitted SMPLX-Lite-D model with the scanned mesh to get chamfer distance (CD, $\times 10^{-3}m$). Std means the average results of 24 standard cameras, and Tele means the average results of 8 telephoto cameras.}
	\label{tab:eval}\\
	
		\toprule
		\multirow{3}{*}{Subject} & \multirow{3}{*}{Act} & \multicolumn{4}{c}{Scan}                            & \multicolumn{4}{c}{SMPLX-Lite-D}                            & \multirow{3}{*}{\begin{tabular}[c]{@{}c@{}}CD$\downarrow$\\($\times 10^{-3}$)\end{tabular}} \\ \cline{3-6} \cline{7-10}
		&                      & \multicolumn{2}{c}{PSNR$\uparrow$} & \multicolumn{2}{c}{SSIM$\uparrow$} & \multicolumn{2}{c}{PSNR$\uparrow$} & \multicolumn{2}{c}{SSIM$\uparrow$} &                                                                                                                   \\ \cline{3-6} \cline{7-10}
		&                      & Std         & Tele       & Std         & Tele       & Std         & Tele       & Std         & Tele       &                                                                                                                   \\ \midrule
		\endfirsthead
		
		\toprule
		\multirow{3}{*}{Subject} & \multirow{3}{*}{Act} & \multicolumn{4}{c}{Scan}                            & \multicolumn{4}{c}{SMPLX-Lite-D}                            & \multirow{3}{*}{\begin{tabular}[c]{@{}c@{}}CD$\downarrow$\\($\times 10^{-3}$)\end{tabular}} \\ \cline{3-6} \cline{7-10}
		&                      & \multicolumn{2}{c}{PSNR$\uparrow$} & \multicolumn{2}{c}{SSIM$\uparrow$} & \multicolumn{2}{c}{PSNR$\uparrow$} & \multicolumn{2}{c}{SSIM$\uparrow$} &                                                                                                                   \\ \cline{3-6} \cline{7-10}
		&                      & Std         & Tele       & Std         & Tele       & Std         & Tele       & Std         & Tele       &                                                                                                                   \\ \midrule
		\endhead
		
		\bottomrule
		\endfoot
		\endlastfoot
		
		\multirow{15}{*}{WZL}    & 01                   & 30.26       & 25.93      & 0.9803      & 0.9425     & \textbf{30.00}       & 25.59      & 0.9797      & 0.9407     & 6.734                                                                                                             \\
		& 02                   & \textbf{30.31}       & 26.00      & \textbf{0.9805}      & 0.9419     & \textbf{30.00}       & 25.47      & \textbf{0.9799}      & 0.9400     & 6.756                                                                                                             \\
		& 03                   & 30.23       & 25.94      & 0.9803      & 0.9415     & 29.83       & 25.32      & 0.9796      & 0.9396     & 6.829                                                                                                             \\
		& 04                   & 30.11       & 25.77      & 0.9800      & 0.9416     & 29.79       & 25.23      & 0.9796      & 0.9400     & 6.534                                                                                                             \\
		& 05                   & 30.05       & 25.59      & 0.9798      & 0.9424     & 29.74       & 25.06      & 0.9792      & 0.9406     & 6.683                                                                                                             \\
		& 06                   & 29.61       & 24.90      & 0.9791      & 0.9412     & 29.36       & 24.51      & 0.9786      & 0.9399     & \textbf{6.465}                                                                                                             \\
		& 07                   & 30.11       & 25.86      & 0.9802      & 0.9448     & 29.80       & 25.47      & 0.9796      & 0.9436     & 6.746                                                                                                             \\
		& 08                   & 29.68       & 24.69      & 0.9790      & 0.9410     & 29.44       & 24.33      & 0.9785      & 0.9398     & 6.644                                                                                                             \\
		& 09                   & 30.07       & 25.77      & 0.9800      & 0.9457     & 29.82       & 25.43      & 0.9795      & 0.9442     & 6.657                                                                                                             \\
		& 10                   & 29.66       & 26.76      & 0.9798      & 0.9567     & 29.43       & 26.43      & 0.9792      & 0.9553     & 6.899                                                                                                             \\
		& 11                   & 29.23       & \textbf{27.12}      & 0.9784      & \textbf{0.9672}     & 29.04       & \textbf{26.83}      & 0.9779      & \textbf{0.9659}     & 7.042                                                                                                             \\
		& 12                   & 29.86       & 25.59      & 0.9797      & 0.9440     & 29.60       & 25.20      & 0.9791      & 0.9425     & 6.648                                                                                                             \\
		& 13                   & 29.86       & 25.71      & 0.9797      & 0.9431     & 29.55       & 25.18      & 0.9791      & 0.9414     & 6.703                                                                                                             \\
		& 14                   & 30.08       & 26.58      & 0.9803      & 0.9561     & 29.82       & 26.19      & 0.9797      & 0.9542     & 6.859                                                                                                             \\
		& 15                   & 29.96       & 25.89      & 0.9796      & 0.9432     & 29.68       & 25.44      & 0.9789      & 0.9416     & 6.858                                                                                                             \\ \midrule
		\multirow{15}{*}{LDF}    & 01                   & 28.72       & 27.03      & 0.9750      & 0.9572     & 28.47       & 26.63      & 0.9743      & 0.9561     & 6.902                                                                                                             \\
		& 02                   & 28.82       & 27.11      & 0.9749      & 0.9566     & 28.59       & 26.70      & 0.9744      & 0.9551     & 6.834                                                                                                             \\
		& 03                   & 28.78       & 27.29      & 0.9747      & 0.9566     & 28.54       & 26.85      & 0.9742      & 0.9553     & 6.852                                                                                                             \\
		& 04                   & 28.83       & 27.13      & 0.9747      & 0.9551     & 28.61       & 26.74      & 0.9742      & 0.9538     & 6.873                                                                                                             \\
		& 05                   & 28.59       & 27.00      & 0.9741      & 0.9548     & 28.35       & 26.57      & 0.9735      & 0.9536     & 7.004                                                                                                             \\
		& 06                   & 28.21       & 26.91      & 0.9734      & 0.9564     & 27.98       & 26.54      & 0.9729      & 0.9553     & 6.882                                                                                                             \\
		& 07                   & 28.72       & 26.92      & 0.9740      & 0.9551     & 28.50       & 26.56      & 0.9735      & 0.9543     & 6.888                                                                                                             \\
		& 08                   & 28.48       & 26.78      & 0.9731      & 0.9536     & 28.21       & 26.31      & 0.9726      & 0.9523     & \textbf{6.840}                                                                                                            \\
		& 09                   & 28.80       & 26.96      & 0.9746      & 0.9576     & 28.58       & 26.59      & 0.9740      & 0.9559     & 7.078                                                                                                             \\
		& 10                   & \textbf{29.15}       & \textbf{28.75}      & \textbf{0.9780}      & \textbf{0.9761}     & \textbf{28.93}       & \textbf{28.46}      & \textbf{0.9773}      & \textbf{0.9751}     & 9.649                                                                                                             \\
		& 11                   & 28.88       & 28.14      & 0.9774      & 0.9710     & 28.65       & 27.82      & 0.9768      & 0.9697     & 9.112                                                                                                             \\
		& 12                   & 28.86       & 27.28      & 0.9745      & 0.9569     & 28.64       & 26.88      & 0.9739      & 0.9556     & 7.215                                                                                                             \\
		& 13                   & 28.54       & 26.98      & 0.9731      & 0.9550     & 27.37       & 25.83      & 0.9655      & 0.9456     & 6.929                                                                                                            \\
		& 14                   & 28.80       & 27.12      & 0.9751      & 0.9606     & 28.61       & 26.79      & 0.9744      & 0.9590     & 7.363                                                                                                             \\
		& 15                   & 28.12       & 26.90      & 0.9721      & 0.9547     & 27.81       & 26.42      & 0.9712      & 0.9532     & 7.198                                                                                                             \\ \midrule
		\multirow{13}{*}{ZX}     & 01                   & \textbf{29.83}       & 27.91      & 0.9799      & 0.9646     & 29.43       & 27.07      & 0.9790      & 0.9623     & 6.885                                                                                                             \\
		& 02                   & 29.52       & 27.56      & 0.9795      & 0.9650     & 29.18       & 26.84      & 0.9786      & 0.9625     & 6.785                                                                                                             \\
		& 03                   & 29.48       & 27.69      & 0.9796      & 0.9664     & 29.20       & 27.09      & 0.9791      & 0.9654     & 6.700                                                                                                             \\
		& 04                   & 29.51       & 27.58      & 0.9791      & 0.9653     & 29.12       & 26.83      & 0.9783      & 0.9629     & 6.633                                                                                                             \\
		& 05                   & 29.36       & 27.52      & 0.9788      & 0.9645     & 28.99       & 26.69      & 0.9779      & 0.9623     & 6.775                                                                                                             \\
		& 06                   & 28.67       & 26.79      & 0.9768      & 0.9621     & 28.37       & 26.09      & 0.9765      & 0.9604     & \textbf{6.514}                                                                                                             \\
		& 07                   & 29.87       & \textbf{28.40}      & 0.9792      & 0.9696     & \textbf{29.56}       & \textbf{27.82}      & 0.9786      & 0.9681     & 6.795                                                                                                             \\
		& 08                   & 28.84       & 27.36      & 0.9778      & 0.9644     & 28.51       & 26.58      & 0.9768      & 0.9618     & 6.768                                                                                                             \\
		& 09                   & 29.23       & 27.44      & 0.9796      & 0.9654     & 28.91       & 26.76      & 0.9789      & 0.9635     & 6.869                                                                                                             \\
		& 10                   & 29.66       & 27.91      & \textbf{0.9806}      & 0.9682     & 29.34       & 27.35      & \textbf{0.9797}      & 0.9663     & 6.911                                                                                                             \\
		& 11                   & 29.37       & 28.19      & 0.9803      & \textbf{0.9728}     & 29.16       & 27.73      & 0.9795      & \textbf{0.9713}     & 7.110                                                                                                             \\
		& 12                   & 29.52       & 27.71      & 0.9794      & 0.9653     & 29.10       & 26.85      & 0.9785      & 0.9628     & 6.825                                                                                                             \\
		& 13                   & 28.94       & 27.91      & 0.9796      & 0.9684     & 28.58       & 27.22      & 0.9788      & 0.9665     & 6.924                                                                                                             \\
		\multirow{2}{*}{ZX}& 14                   & 29.38       & 27.69      & 0.9800      & 0.9668     & 29.10       & 27.09      & 0.9791      & 0.9645     & 6.941                                                                                                             \\
		& 15                   & 29.42       & 27.75      & 0.9788      & 0.9648     & 28.95       & 26.76      & 0.9776      & 0.9622     & 6.917                                                                                                             \\ \midrule
	
		\multirow{15}{*}{LW}     & 01                   & 28.05       & 25.45      & 0.9799      & 0.9625     & 27.70       & 24.67      & 0.9794      & 0.9612     & 6.312                                                                                                             \\
		& 02                   &     \textbf{28.49}       &   \textbf{26.70}         &  0.9775           &   0.9657         &    \textbf{28.12}         &     \textbf{26.31}       &  0.9770           &       0.9646     &   6.192                                                                                                                \\
		& 03                   &  28.29           &     26.24       &   0.9768          &   0.9652         &     27.90        &     25.82       &  0.9763           &   0.9642         &       6.254                                                                                                            \\
		& 04                   & 28.08       & 25.68      & 0.9800      & 0.9602     & 27.67       & 24.67      & 0.9791      & 0.9579     & 6.379                                                                                                             \\
		& 05                   & 28.07       & 25.38      & 0.9798      & 0.9603     & 27.59       & 24.40      & 0.9790      & 0.9581     & 6.394                                                                                                             \\
		& 06                   & 27.45       & 25.10      & 0.9786      & 0.9577     & 27.07       & 24.20      & 0.9780      & 0.9558     & \textbf{6.184}                                                                                                             \\
		& 07                   & 27.49       & 25.27      & 0.9798      & 0.9644     & 27.10       & 24.50      & 0.9789      & 0.9627     & 6.429                                                                                                             \\
		& 08                   & 27.96       & 25.26      & 0.9789      & 0.9560     & 27.44       & 24.22      & 0.9779      & 0.9540     & 6.488                                                                                                             \\
		& 09                   & 27.99       & 25.20      & \textbf{0.9806}      & 0.9630     & 27.52       & 24.25      & 0.9797      & 0.9608     & 6.622                                                                                                             \\
		& 10                   & 27.82       & 25.05      & 0.9806      & 0.9654     & 27.42       & 24.24      & \textbf{0.9798}      & 0.9635     & 6.517                                                                                                             \\
		& 11                   & 28.07       & 25.93      & 0.9800      & \textbf{0.9705}     & 27.70       & 25.24      & 0.9792      & \textbf{0.9692}     & 6.691                                                                                                             \\
		& 12                   & 28.31       & 25.47      & 0.9798      & 0.9631     & 27.86       & 24.74      & 0.9788      & 0.9614     & 6.609                                                                                                             \\
		& 13                   & 27.32       & 24.54      & 0.9796      & 0.9649     & 26.83       & 23.89      & 0.9788      & 0.9633     & 6.282                                                                                                             \\
		& 14                   & 27.75       & 25.00      & 0.9805      & 0.9673     & 27.33       & 24.18      & 0.9797      & 0.9656     & 6.478                                                                                                             \\
		& 15                   & 27.71       & 25.33      & 0.9790      & 0.9572     & 27.21       & 24.25      & 0.9782      & 0.9551     & 6.316   \\ \midrule
			\multirow{15}{*}{ZC}     & 01                   & \textbf{29.01}       & 25.78      & 0.9712      & 0.9358     & \textbf{28.61}       & 25.07      & 0.9704      & 0.9344     & 6.958                                                                                                             \\
		& 02                   & 28.04       & 25.43      & 0.9705      & 0.9367     & 27.67       & 24.81      & 0.9698      & 0.9357     & 7.033                                                                                                             \\
		& 03                   & 27.79       & 25.69      & 0.9706      & 0.9378     & 27.61       & 25.14      & 0.9699      & 0.9368     & 7.038                                                                                                             \\
		& 04                   & 27.63       & 25.69      & 0.9704      & 0.9378     & 27.40       & 25.13      & 0.9696      & 0.9372     & 7.110                                                                                                             \\
		& 05                   & 27.94       & 25.19      & 0.9672      & 0.9350     & 27.55       & 24.56      & 0.9663      & 0.9342     & 6.907                                                                                                             \\
		& 06                   & 28.02       & 25.15      & 0.9691      & 0.9286     & 27.42       & 24.33      & 0.9640      & 0.9181     & 6.706                                                                                                             \\
		& 07                   & 28.16       & 25.51      & 0.9710      & 0.9371     & 27.40       & 24.62      & 0.9610      & 0.9158     & \textbf{6.404}                                                                                                             \\
		& 08                   & 28.40       & 25.35      & 0.9698      & 0.9299     & 27.97       & 24.62      & 0.9689      & 0.9290     & 6.862                                                                                                             \\
		& 09                   & 28.35       & 25.47      & 0.9706      & 0.9341     & 28.06       & 24.85      & 0.9698      & 0.9333     & 7.099                                                                                                             \\
		& 10                   &  28.64           & \textbf{26.70}           &  0.9724           &  \textbf{0.9570}          &     28.28        &    \textbf{26.35}        &   0.9718          &   \textbf{0.9561}         &     6.684                                                                                                              \\
		& 11                   & 28.23       & 25.32      & 0.9722      & 0.9415     & 27.93       & 24.78      & 0.9714      & 0.9408     & 6.973                                                                                                             \\
		& 12                   & 28.43       & 25.69      & 0.9710      & 0.9380     & 28.07       & 25.05      & 0.9700      & 0.9373     & 7.132                                                                                                             \\
		& 13                   & 27.99       & 25.38      & 0.9698      & 0.9464     & 27.52       & 24.69      & 0.9688      & 0.9448     & 6.665                                                                                                             \\
		& 14                   & 28.56       & 25.46      & \textbf{0.9750}      & 0.9515     & 28.19       & 24.89      & \textbf{0.9740}      & 0.9501     & 7.033                                                                                                             \\
		& 15                   & 28.16       & 25.37      & 0.9699      & 0.9356     & 27.76       & 24.72      & 0.9689      & 0.9344     & 7.013                                                                                                             
		\\ \bottomrule   
\end{longtable}
\clearpage
\twocolumn

\end{document}